\documentclass[sigconf]{acmart}

\AtBeginDocument{%
  \providecommand\BibTeX{{%
    \normalfont B\kern-0.5em{\scshape i\kern-0.25em b}\kern-0.8em\TeX}}}

\copyrightyear{2024}
\acmYear{2024}
\setcopyright{rightsretained}
\acmConference[KDD '24]{Proceedings of the 30th ACM SIGKDD Conference on
Knowledge Discovery and Data Mining}{August 25--29, 2024}{Barcelona, Spain}
\acmBooktitle{Proceedings of the 30th ACM SIGKDD Conference on Knowledge
Discovery and Data Mining (KDD '24), August 25--29, 2024, Barcelona,
Spain}\acmDOI{10.1145/3637528.3672065}
\acmISBN{979-8-4007-0490-1/24/08}

\usepackage{multirow} % use multirow in table
\usepackage{makecell} % use makecell in table

\usepackage{longtable} % use long table
\usepackage{CJKutf8} % to use Chinese with pdfLatex (only supported engine by Arxiv)

\usepackage{tablefootnote}

\usepackage{pifont} % cross and tick
\newcommand{\cmark}{\ding{51}} % cross and tick
\newcommand{\xmark}{\ding{55}} % cross and tick

\usepackage{amsmath,amsfonts,bm}

\def\eqref#1{equation~\ref{#1}}
\def\1{\bm{1}}

\def\va{{\bm{a}}}

\def\vr{{\bm{r}}}

\DeclareMathAlphabet{\mathsfit}{\encodingdefault}{\sfdefault}{m}{sl}
\SetMathAlphabet{\mathsfit}{bold}{\encodingdefault}{\sfdefault}{bx}{n}

\def\gD{{\mathcal{D}}}

\def\sD{{\mathbb{D}}}
\newcommand{\E}{\mathbb{E}}

\usepackage{array}
\newcolumntype{C}[1]{>{\centering\let\newline\\\arraybackslash\hspace{0pt}}m{#1}}
\usepackage[shortcuts]{extdash} % allow line break for words with hypen \-/. https://stackoverflow.com/questions/2193307/how-do-i-get-latex-to-hyphenate-a-word-that-contains-a-dash

\begin{document}

%%
%% The "title" command has an optional parameter,
%% allowing the author to define a "short title" to be used in page headers.
% \title{FoRAG: A Bilingual Retrieval Augmented Generation Model with Fine-grained Factuality Optimization} % RAG is a framework. We propose a new one. we also propose a new model.

\title {FoRAG: Factuality-optimized Retrieval Augmented Generation for Web-enhanced Long-form Question Answering }
% \title{FoRAG: Retrieve, Generation and Fine-grained Factuality Optimization in Bilingual}
% \title{FoRAG: A Bilingual Model with Fine-grained Factuality Optimization}
% \title{FoRAG: Optimizing Factuality in a Fine-grained Manner} % 容易和 summarization 工作 overlap？(FoRAG: Factuality Optimized RAG ) 
 % FactRAG, 或者是否要替换掉 factuality consistency 比如用 truthful 或者 faithful, FaithfulRAG: FaithRAG 
%\title{Bilingual Factuality Optimized RAG} % 关键词太多有点难记
% \title{A fine-grained factuality optimized bilingual RAG.}

%%
%% The "author" command and its associated commands are used to define
%% the authors and their affiliations.
%% Of note is the shared affiliation of the first two authors, and the
%% "authornote" and "authornotemark" commands
%% used to denote shared contribution to the research.
\author{Tianchi Cai}
\affiliation{%
  \institution{Ant Group}
  \city{Hangzhou}
  \country{China}
}
\email{tianchicai@gmail.com}

\author{Zhiwen Tan}
\affiliation{%
  \institution{Ant Group}
  \city{Hangzhou}
  \country{China}
}
\email{ender.tzw@antgroup.com}

\author{Xierui Song}
\affiliation{%
  \institution{Ant Group}
  \city{Hangzhou}
  \country{China}
}
\email{songxierui.sxr@antgroup.com}

\author{Tao Sun}
\affiliation{%
  \institution{Ant Group}
  \city{Hangzhou}
  \country{China}
}
\email{suntao.sun@antgroup.com}

\author{Jiyan Jiang}
\affiliation{%
  \institution{Tsinghua University}
  \city{Beijing}
  \country{China}
}
\email{scjjy95@outlook.com}

\author{Yunqi Xu}
\affiliation{%
  \institution{Ant Group}
  \city{Hangzhou}
  \country{China}
}
\email{xuyunqi.xyq@antgroup.com}
  
\author{Yinger Zhang}
\affiliation{%
  \institution{Ant Group}
  \city{Hangzhou}
  \country{China}
}
\email{zhangyinger@zju.edu.cn}

\author{Jinjie Gu}
\affiliation{%
  \institution{Ant Group}
  \city{Hangzhou}
  \country{China}
}
\email{jinjie.gujj@antgroup.com}

%%
%% By default, the full list of authors will be used in the page
%% headers. Often, this list is too long, and will overlap
%% other information printed in the page headers. This command allows
%% the author to define a more concise list
%% of authors' names for this purpose.
\renewcommand{\shortauthors}{Cai and Tan, et al.}

%%
%% The abstract is a short summary of the work to be presented in the
%% article.
\begin{abstract}

Retrieval Augmented Generation (RAG) has become prevalent in question-answering (QA) tasks due to its ability of utilizing search engine to enhance the quality of long-form question-answering (LFQA). Despite the emergence of various open source methods and web-enhanced commercial systems such as Bing Chat, two critical problems remain unsolved, i.e., the lack of factuality and clear logic in the generated long-form answers. In this paper, we remedy these issues via a systematic study on answer generation in web-enhanced LFQA. Specifically, we first propose a novel outline-enhanced generator to achieve clear logic in the generation of multifaceted answers and construct two datasets accordingly. Then we propose a factuality optimization method based on a carefully designed doubly fine-grained RLHF framework, which contains automatic evaluation and reward modeling in different levels of granularity. Our generic framework comprises conventional fine-grained RLHF methods as special cases. Extensive experiments verify the superiority of our proposed \textit{Factuality-optimized RAG (FoRAG)} method on both English and Chinese benchmarks. In particular, when applying our method to Llama2-7B-chat, the derived model FoRAG-L-7B outperforms WebGPT-175B in terms of three commonly used metrics (i.e., coherence, helpfulness, and factuality), while the number of parameters is much smaller (only 1/24 of that of WebGPT-175B). Our datasets and models are made publicly available for better reproducibility.\footnote{\url{https://huggingface.co/forag} \label{footnote_dataset_url}}
%, such as Bing Chat, ensuring the factual integrity of their outputs is paramount. Addressing this challenge, we introduce a pioneering factuality-optimized RAG model. Our contributions are threefold. To improve answer quality in terms of helpfulness and coherence, we compiled a dataset comprised of diverse real-world user queries in Chinese and English, paired with references from large-scale search engines. We further introduce an two-stage outline-enhanced generator technique. Initially, it constructs a structured answer outline to ensure logical progression, followed by detailed content generation that adheres to the established outline, thus fostering coherence. Moreover, we propose an innovative automatic factuality optimization framework. This framework is the first to systematically address truthfulness at multiple granular levels, from the precision of fact-checking to the intricacies of reward modeling in the optimization process. This framework generalizes and extends beyond several existing methods in fined-grained Reinforcement Learning from Human Feedback (RLHF), presenting a suite of fine-grained techniques for improving factuality. Our comprehensive experiments demonstrate the efficacy of our methods. When applied to Llama2-7B parameters, our model outperforms other much larger models, such as WebGPT-175B. The results highlight the efficacy of our approach in generating more factual and organized responses, underscoring the potential of fine-grained optimization in the domain of web-scale RAG applications. \footnote{\url{https://huggingface.co/forag}}
\end{abstract}

%%
%% The code below is generated by the tool at http://dl.acm.org/ccs.cfm.
%% Please copy and paste the code instead of the example below.
% %%

\maketitle

\begin{figure*}[t] 
  \centering
  \includegraphics[width=\linewidth]{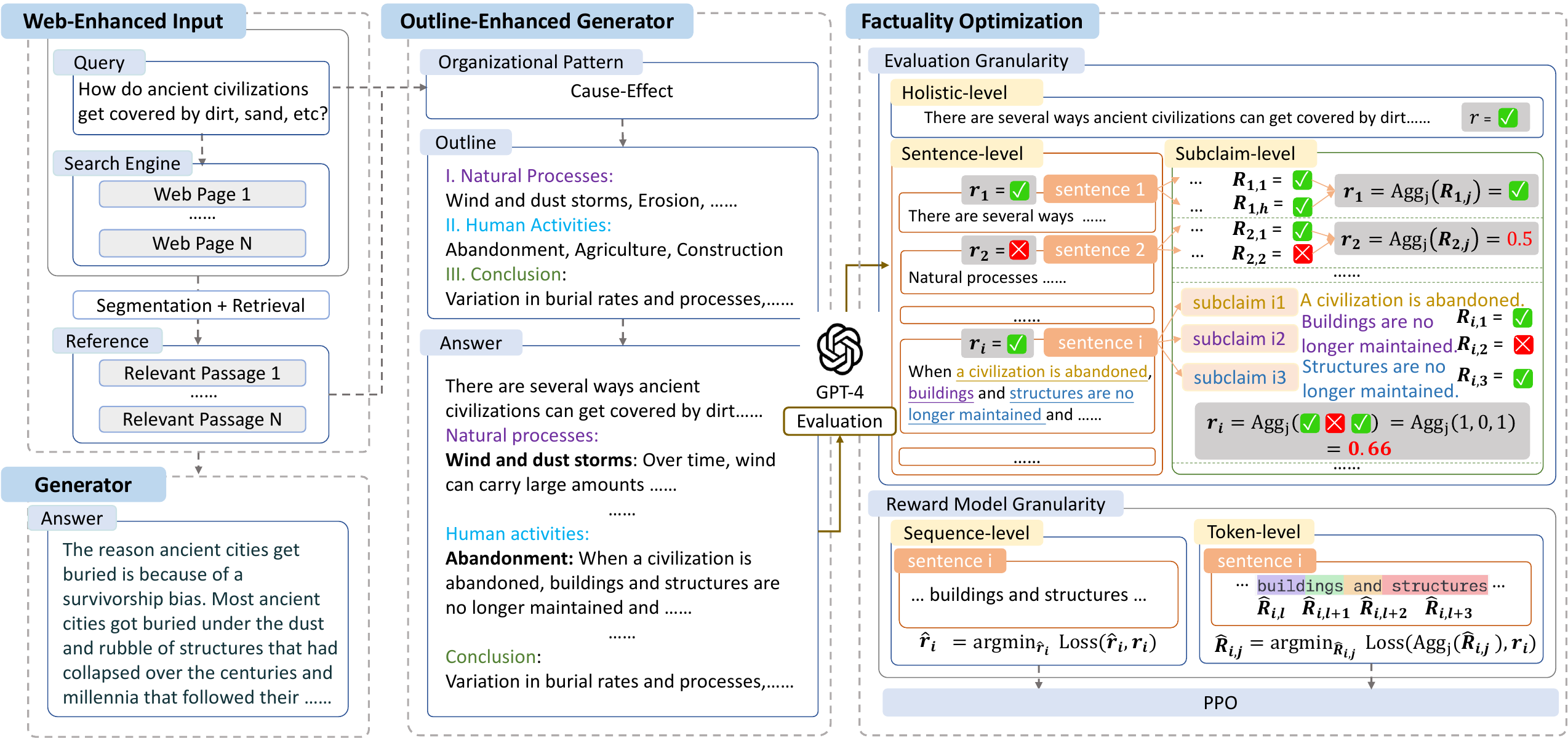}
  \caption{Illustrations of the input for LLM in web-enhanced LFQA task (upper left), the existing generator (lower left), our outline-enhanced generator  (middle) and our doubly fine-grained factuality optimization method (right). Before generating a long answer, the outline-enhanced generator first drafts an organization pattern and  an outline to promote a clear logic for generation. The doubly fined-grained RLHF optimizes factuality by incorporating fine-grained designs on two core steps, i.e. factuality evaluation and reward modeling, with methods on multiple levels of granularities proposed on each step. } \label{caption}
\end{figure*}

\section{Introduction}

Retrieval Augmented Generation (RAG), a technique that augments Large Language Models (LLMs) with a retriever by appending the retrieved relevant passages to the current context \cite{mialon2023augmented}, has recently attracted considerable research attention \cite{borgeaud2022improving,guu2020retrieval}. The access to search engine supplements massive and latest knowledge to LLMs, boosting their performance on various knowledge intensive tasks, such as open domain dialogue \cite{thoppilan2022lamda} and question answering (QA) \cite{shuster2022blenderbot}. 

Following this paradigm, many web-enhanced commercial systems have been developed, such as Bing Chat\footnote{Bing Chat: \url{https://www.bing.com/chat}. \label{bingchat}}  and perplexity.ai.\footnote{perplexity.ai: \url{https://www.perplexity.ai}.} They generate answers to user queries in natural language with reference to web pages, which we refer to as the web-enhanced long-form question-answering (LFQA) task. Although these systems can generate coherent and helpful answers, recent researches have revealed the low factuality issue of these systems, such that only about half of the statements generated are fully supported by the retrieved references \cite{liu2023evaluating,gao2023enabling}. This poses an unignorable threat to the trustworthiness of existing web-enhanced LFQA systems.

%Recently multiple researches independently reveal that though these cutting-edge web-enhanced LFQA systems in general provides coherent and helpful answers, only about half of the statements generated are fully supported by the corresponding references \cite{liu2023evaluating,gao2023enabling}. This unsatisfyingly low factuality posed a significant threaten to the trustiness between potential knowledge seeking users and these web-enhanced LFQA systems.

Despite its prevalence, there lacks an effective method to optimize factuality in web-enhanced LFQA as far as we are concerned. There are two intrinsic difficulties. First, previous studies mostly rely on human evaluation \cite{nakano2021webgpt,menick2022teaching,liu2023evaluating},  which is generally expensive to acquire. In web-enhanced LFQA task, factuality is even more time-consuming and difficult to manually annotate compared to other language generation tasks, since it involves comparing the factual details of two lengthy texts \cite{wu2023fine}. Second, the most commonly used fine-tuning method for human preference alignment, i.e., Reinforcement Learning from Human Feedback (RLHF), conventionally adopts the \textit{holistic} reward, such that each answer only has a single evaluation score. Such a reward provides a relatively sparse training signal, which undermines the reliability of RLHF \cite{ramamurthy2022reinforcement,wu2023fine}. In web-enhanced LFQA, the sparsity problem is even exaggerated, as the answers are in long form. 

Besides the above factuality issue, different from conventional QA tasks with short answers, web-enhanced LFQA poses extra challenges due to the pervasive ambiguity of many real-world questions. A desirable answer to these questions is preferred to be multifaceted \cite{amplayo2022query}, which requires organizing and  consolidating information from multiple aspects and references \cite{kulkarni2020aquamuse}. The problem might be one possible reason why existing open source methods such as WebGLM \cite{liu2023webglm} have no explicit improvement over closed source methods such as WebGPT\-/175B.

To resolve the above issues, in this paper, we conduct a systematic study of web-enhanced LFQA. Specifically, we first propose a novel outline-enhanced generator, which achieves clear logic in the generation of multifaceted answers. We then propose an innovative factuality optimization approach based on a novel doubly fine-grained RLHF framework. Specifically, we design new automatic evaluation and reward modeling steps in different granularities, which allows to optimize factuality for RAG in a flexible way. Our generic method contains several existing fine-grained RLHF methods as special cases. 

% Specifically, our proposed approach consists of two modules, i.e., automatic evaluation and reward modeling. To overcome the sparse reward issue of traditional holistic reward, both modules in our approach

% Unlike previous approach conducted in solely holistic granularity, both modules are designed for various granularities, which allows to optimize factuality for RAG in a finer-grained way. 

%Note that although the doubly fine-grained RLHF is proposed for factuality optimization for RAG, it can also be generalized to other metrics as well as other language generation tasks. 

%Our \textit{Factuality-optimized RAG (FoRAG)} contains two novelly designed modules, a fine-grained automatic factuality evaluation module and a fine-grained reward model module. \TODO{different granularity combination} As far as we know, our factuality optimization technique is the first that tackles the factuality problem in web-enhanced RAG task. Note that though the methods is designed for optimizing factuality, it provides a general framework for fine-grained RLHF, which generalize several existing fine-grained RLHF techniques as special cases, and can be applied to optimize other metrics easily.

Extensive experiments demonstrate the effectiveness of our proposed method, which achieves state-of-the-art performance on both Chinese and English benchmarks. Specifically, the outline-enhanced generator significantly improves the coherence and helpfulness, while the doubly fine-grained factuality optimization method substantially promotes the factuality on both answer and sentence levels. Remarkably, applying our method to Llama2-7B-chat yields a fine-tuned model FoRAG-L-7B, which, for the first time, surpasses the WebGPT\-/175B on coherence, helpfulness, and factuality, while the number of parameters of FoRAG-L-7B is much smaller (only 1/24 of that of WebGPT\-/175B).

%\TODO{the first time} Notablily, applying our methods on Llama2-7B-chat yields a finetuned model, i.e., FoRAG-L 7B, that surpasses the WebGPT\-/175B (even in ... size), which contains 24 times more parameters than our model. \TODO{method dataset are open-source}

The contributions of this work are summarized as follows:
\begin{itemize}
    \item We propose a new outline-enhanced generator to promote a clear logic of long answer generation in RAG, which significantly improves the coherence and helpfulness of the generated answers. Two large-scale outline-enhanced LFQA datasets are accordingly constructed. 
    \item We propose a novel factual optimization method for web-enhanced RAG based on a novel doubly fine-grained RLHF framework, which eschews the need of expensive human annotation. %The proposed method can be generalized to other factuality optimization problems.
    \item We conduct extensive experiments to show that our method achieves state-of-the-art performance on both Chinese and English benchmarks. Notably, the FoRAG-L-7B model fine\-/tuned by our method outperforms WebGPT-175B on coherence, helpfulness, and factuality, with only 1/24 of the parameters in WebGPT-175B. Both datasets and models are publicly available for better reproducibility.  
 \end{itemize}

%\end{todo}

% Generative large language models (LLMs) \cite{openai2023gpt,anthropic2023model,touvron2023llama} often struggle to provide factually accurate statemnets, resulting in the hallucination problem \cite{ji2023survey}. Recent efforts tackling this issue by augmenting the LLMs with external tools  \cite{mialon2023augmented} such as retrievers \cite{shuster2021retrieval,lewis2020retrieval,guu2020retrieval} or search engines \cite{nakano2021webgpt,thoppilan2022lamda,qin-etal-2023-webcpm,liu2023webglm}. Incorporating external references backs the generated statements with the references, and enables the evaluation of factuality, that is, whether the generated statement is fully verifiable by the reference. 

\section{Related Work}
In this section, we review prior work in three related fields, i.e., open-domain question answering, retrieval augmented generation, as well as web-enhanced LFQA.

\textbf{Open-domain Question Answering}. 
In the field of open-domain QA, traditional efforts have centered around reading comprehension techniques, with foundational datasets like SQuAD \cite{rajpurkar2016squad} providing human-written questions and extracted answers. Subsequent datasets, including Natural Questions \cite{kwiatkowski2019natural}, TriviaQA \cite{joshi2017triviaqa}, and CoQA \cite{reddy2019coqa}, continue this trend but largely cater to brief answers. Recognizing the value of more informative, long-form responses, recent initiatives such as ELI5 \cite{eli5_lfqa} have begun to compile questions demanding detailed answers, prompting research into advanced generative techniques to meet this need.

\textbf{Retrieval-Augmented Generation}.
Retrieval-Augmented Generation (RAG) enhances language model (LM) performance by integrating external knowledge retrieval with in-context learning. The knowledge retrieval techniques in RAG include sparse methods such as BM25 and TF-IDF and dense retrieval systems, including DPR \cite{karpukhin2020dense} and Contriever \cite{izacard2021unsupervised}. To utilize the retriever, various methods are proposed. REALM proposes a joint optimization of retriever and language modeling \cite{guu2020retrieval}. Retro uses a frozen retriever to enhance the generation ability model with a novel chunked cross-attention mechanism \cite{borgeaud2022improving}. Atlas studies the few-shot ability for RAG models. Others combine black-box LMs with custom or fine-tuned retrieval systems~\citep{shi2023replug,ram-etal-2023-context}. Different from these work, we treat the retrieval step as a black box and focus on improving the generation quality given the query and retrieved passages.

\textbf{Web-enhanced LFQA}.
The web-enhanced LFQA takes a new approach to QA tasks which utilizes the retrieval ability of search engine to improve the generation performance on long-form QA. Closely related to our work, WebGPT~\cite{nakano2021webgpt} uses the questions from ELI5 and explores the ability of LLMs in navigating through the web, retrieving passages from web, and generating long-form responses. Despite its notable capabilities, the dataset, and models are not accessible to the public. Following this idea, WebCPM~\cite{qin-etal-2023-webcpm} builds and releases an open-source web-enhanced LFQA model in Chinese. WebGLM~\cite{liu2023webglm} provides an more efficient and cost-effective method by replacing the expert annotation with evaluations using LLMs and utilizing a non-interactive way to use search engine. However, its resulting model, the WebGLM-10B does not outperforms the WebGPT-175B. Compared to these works, we consider optimizing the logic structure and factuality of the generation, which has not been studied in web-enhanced LFQA as far as we know.

\section{Preliminary}
In this section, we briefly review the  RAG pipeline in web-enhanced LFQA task, which for simplicity of presentation, we adopt the term \textit{web-enhanced RAG} to describe in the sequel. The web-enhanced RAG pipeline is demonstrated in the left column of Figure \ref{caption}.

In web-enhanced RAG, for a given user input $x$, the system first utilizes a web search engine to retrieve a list of relevant website URLs, then crawls the websites and extracts the relevant text segments $z$, which are usually called \textit{reference} or \textit{context} for generation \cite{qin-etal-2023-webcpm}. This extraction is commonly done by first segmenting the web pages into text segments and then using pre-trained dense retrievers to extract the top-k segments \cite{liu2023webglm}.

After deriving the context $z$, the RAG system generates an answer $y$ based on the context $z$ and the user query $x$. Following \cite{ramamurthy2022reinforcement}, the response generation can be formulated as a Markov Decision Process (MDP) $<\mathcal{S}, \mathcal{A}, \mathcal{R}, P, \gamma>$. In such a process, each episode starts with a sampled state $s \in \mathcal{S}$, where $s=(x,z)$ is a prompt that contains a query $x$ and a relevant context $z$ (here the parenthesis denotes string concatenation). At each step $t$ during this episode, the state $s_t=(x,z,a_1,...,a_{t-1})$ is described by the query $x$, the context $z$, and all the previously generated tokens $(a_1,...,a_{t-1})$, {which is denoted $\boldsymbol{a}_{t-1}$ for short}. Given the state $s_t$, the LLM, denoted by $\pi_\theta$, defines a probability distribution  $a \sim \pi_\theta(\cdot|s_t)$ over all tokens $a\in\mathcal{A}$ conditioned on the current state $s_t$, where $\theta$ denotes the trainable parameters of the LLM. After generating the specific token $a_t\in\mathcal {A}$, the state will transit to $s_{t+1}=(s_t,a_t)$ at the next time step $t+1$ by appending the latest generated $a_t$ token to the current state $s_t$. This episode terminates when the length $t$ exceeds a pre-defined threshold $T$ or an end-of-sequence token is generated. In the above definition of MDP, the parameter $\gamma$ is a discount factor. In most of the language generation tasks, we have a task specific evaluation metric $\mathcal{R} (\va_T, x, z) $ that depends on the final context $s_T = (x,z,a_1,...,a_T)$ which consists of the generated sequence $\va_T$ and the initial context $x,z$, which is typically given at the end of sequence to reflect the quality of the generated sequence, e.g., whether the sequence is helpful or harmless \cite{bai2022training}. Depending on the evaluation granularity, $\mathcal{R} (\va_T, x, z) $ might be a scalar or a vector or even a matrix, we denotes the three cases by $r$, $\vr$, $\boldsymbol{R}$, respectively (see later explanations in Section \ref{evaluation}) .

Besides helpfulness or harmlessness, one crucial criterion of response generation in RAG is factuality (or verifiability), which leverages the extent to which the generated response is trustful. In general, the response $y$ is considered to be truthful if its contents are factually consistent with the retrieved text $z$ \cite{wang2023survey}. In most previous works \cite{min2023factscore,wang2023evaluating}, factuality is mainly evaluated by human annotation or via the NLI model (i.e., whether the context can entail the information contained by the response) \cite{zha2023alignscore} or general purpose LLMs \cite{kamoi2023wice,lattimer2023fast,hu-etal-2023-refchecker}, such as ChatGPT \cite{ouyang2022training} or GPT4 \cite{openai2023gpt}.

% 参考 xx 工作，在 RAG 中，我们使用 response 和 context 里的 factual information 是否一致作为 factuality 的指标。现有的工作主要都是使用人工标注，或者 NLI 模型预估 context 是否可以 entail response 中的信息。我们也使用了相同的定义方式（见 sec \ref{4.3})

% \section{Factuality-Optimized RAG}
\section{Outline-Enhanced RAG}

% In this section, we propose the \textit{Factuality-Optimized RAG (FoRAG)} technique, which is able to generate well-structured and factual responses of high quality. FoRAG has two novel designs, i.e., a two-stage generator with outline-enhanced technique to generate answers with organizational patterns, and a multi-granularity factuality optimization module to promote the factuality of the responses. In the following, we will describe our novel designs in detail. The detailed demonstration of the FoRAG is presented in Figure \ref{caption} \TODO{(middle and right)}.

{In this section, we propose the \textit{Outline-Enhanced} generation technique, which is able to generate well-structured responses of high quality. Illustrated in Figure \ref{caption}} (middle), the outline-enhanced generator takes a two-stage generation, where the generator first generates an organizational pattern and outline to improve the logic structure. In the following, we describe our technique and the corresponding construction of two outline-enhanced LFQA datasets. 

\subsection{Outline-Enhanced Generator}
% \paragraph{Retriever} Though it has been show that when using few-shot examples, LLMs can treat the retrieval task as a generation task, and such generative retrievers outperforms the classic dense retrievers, this approach is not widely adopted due to the slow inference speed \cite{}. With the help of a recently inference acceleration techniques, which speeds up the inference time of LLM on such generative retrieving task by 5.36 times \cite{zhao2023lookahead}, we adopt the same acceleration techniques and trained a LLM to retrieve relevant segments from the web sites. After we trained an LLM, at inference, each website $z$ is concatenated with the user query $x$ and feeded into the model, to generate segments that are relevant to answer the question. If the web site is not helpful to answer the quesiton, the LLM simply gives empty output. All web sites are processing in parallel to speed up the process. 

% The main idea is to build a prefix tree using the context as well as some pre-defined datasets. The inference is then replaced by an alternative between rollout several possible generations using the prefix tree, then verify all these possibilities in one forward pass. The attention mask is carefully designed to allow verifying multiple possibilities in one pass. Then the longest correct subsequence is used as the generation and the method continue of rollout based on this sequence. We include a full description of the method in Appendix \cite{} to make our article self-contained. 

In most existing open-source methods \cite{liu2023webglm,qin-etal-2023-webcpm,nakano2021webgpt}, the responses are directly generated, i.e., the retrieved contents are concatenated with the original query and fed into a generation model using certain prompt template (Figure \ref{caption} lower left). Compared to those generated by closed-source methods, these responses are shorter and often found unorganized, lacking a clear logical structure.

\begin{table}[t]
\centering
\caption{{The statistics of web-enhanced long-form QA datasets. Applying our outline-enhanced generation technique yields significantly longer answers (Ans. (OE)) compared with the original answer (Ans. (Ori)).}}
\label{tab:dataset}
\begin{tabular}{@{}ccccc@{}}
\toprule
\multirow{2}{*}{\textbf{Dataset}}  & \multicolumn{3}{c}{\textbf{Avg. Length (in Tokens)}\tablefootnote{We utilize the tokenizer of \url{https://huggingface.co/meta-llama/Llama-2-7b-chat-hf}.}} & \multirow{2}{*}{\textbf{\makecell{\# Samples}}} \\ \cmidrule(lr){2-4}
 &  Query & Ans. (Ori.) & Ans. (OE) \\ \midrule
% following is from ChatGLM2
% WebCPM (zh) & 18.1 & 159.7 & 250.2 & 5,500 \\ \midrule
% WebGLM (en) & 16.5 & 139.8 & 368.0 & 43,579 \\ \midrule
% WebGPT-13b (en) & 19.4 & 196.3 & 366.7 & 272 \\ \midrule
% WebGPT-175b (en) & 19.4 & 193.7 & 372.8 & 272 \\
%% following is from LLaMA2
WebCPM (zh) & 44.1 & 374.1 & 623.1 & 5,500 \\ \midrule
WebGLM (en) & 17.8 & 151.3 & 409.0 & 43,579 \\ \midrule
WebGPT-13b (en) & 20.9 & 212.4 & 407.4 & 272 \\ \midrule
WebGPT-175b (en) & 20.9 & 208.9 & 414.2 & 272 \\
\bottomrule
\end{tabular}%
\vspace{-2mm}
\end{table}

To enhance the performance, one possible way is to make the responses more organized. Indeed, some researchers have found that carefully designed prompts that comprise task descriptions and a few demonstrations will improve the quality of the generated responses on various tasks \cite{brown2020language}. For example, the technique of ``Let's think step by step" \cite{kojima2022large} substantially improves the performance by encouraging the chain-of-thought reasoning ability. 

%Existing open source work is directly generated. We found that these answers often lacked a clear logical structure. But by way of comparison, the answers generated by closed-source work tend to have a clear logical structure. So we want to enhance the work we already have to make these answers more logical, which in turn may improve the results

%The retrieved contents are then concatenated with the original query and feed into a generation model using certain prompt template. We find that closed-source productions usually generate contents with a clear logic structure and format, which is not presented in existing open-source method. 

%Researchers have found that design prompts comprising task descriptions and a few demonstrations improves the model performance on various tasks \cite{}. There have been various techniques to improve the generation quality using prompting techniques, such as using ``Let's think step by step" \cite{kojima2022large} will encourage the Chain-of-Thought reasoning ability. 

Inspired by the above works, we introduce the outline-enhanced technique into response generation. Our proposed generator includes an outline stage and an expansion stage, which aligns with the intuition that when answering a question, human usually first outlines and organizes the answer before expanding each point. Specifically, to generate high-quality output with a clear logic flow, we prompt the model to first output an outline of the final answer, and then concatenate the draft into the prompt to generate the full response. In the following, we elaborate the two stages in detail.

% \ref{retrieval_prompts}
% \ref{generation_prompts_ch}
% \ref{generation_prompts_en}

\textbf{Outline Stage.} In this stage, the generator first drafts an outline of the answer using an outline template, with the user query $x$ and context $z$ as input. The outline template guides the LLM to first consider which organizational pattern is best suitable to the current question, e.g.,  ``cause and effect" or ``compare and contrast". Then the LLM uses the organizational pattern to output an outline. For example, when the selected pattern is ``compare and contrast", the generated outline will include various perspectives that will later be used to expand on the similarities and differences.

\textbf{Expansion Stage.} Based on the outline generated at the former stage, the LLM expands each perspective to construct the final answer. Specifically, The model is then asked to generate an answer to the question, given the input containing the query $x$, the context $z$ and the outline $o$ generated in the first stage.

The training of the outline-enhanced generator follows the standard supervised fine-tuning (SFT) procedure, which is widely adopted in previous works \cite{ouyang2022training,bai2022training}. 

%We then describe the construction of two outline-enhanced  datasets we used for training our outline-enhanced generator.

\subsection{Outline-Enhanced Long-Form QA Dataset}

{As far as we know, there are only two open-sourced web-enhanced long-form QA datasets available for training web-enhanced RAG models.\footnote{The WebGPT demo website contains 272 samples, which can be used for evaluation but not sufficient for training.} The English dataset, i.e. the WebGLM-QA  \cite{liu2023webglm}, contains $44k$ samples, while the Chinese dataset, i.e. WebCPM \cite{qin-etal-2023-webcpm}, contains $5,500$ samples. The queries in both datasets are sampled from ELI5~\cite{eli5_lfqa}, where WebGLM-QA sample question from it, and WebCPM additionally uses human annotators to translate the question into Chinese. The Web search engine are used to collect relevant passages.}

We construct an outline-enhanced bilingual long-form QA dataset using the queries and relevant passages from these two datasets. We apply our outline-enhanced generation technique using GPT4 \cite{achiam2023gpt} to collect outline-enhanced answers. We design a prompt to instruct GPT4 to execute the outline stage and the expansion stage in a step-by-step manner, which is provided in Appendix~\ref{app:prompt_geneartion}.
%Our preliminary analysis shows that GPT4 is capable of generating high-quality outline and answers in one step, while utilizing a two step approach leads to more hallucinations. Therefore, when using GPT4, we merge the instructions of outline stage and expansion stage into one prompt to construct the outline and answers ({the prompt is provided in Appendix~\ref{app:prompt_geneartion}}).
The detailed statistics of the existing datasets and our outline-enhanced answers are presented in Table \ref{tab:dataset}. It is clear that our demonstration answers are much longer than that in existing works, due to the stronger logic structure (examples are provided in Appendix \ref{dataset_answer_improvement}). 
The outline-enhanced answers derived from WebCPM and WebGLM are publicly available. \textsuperscript{\ref{footnote_dataset_url}}

Note that the imbalance of the amount of training samples in English and Chinese datasets may pose difficulty to train a bilingual web-enhanced RAG model. To overcome this difficulty, we further collect $39k$ queries and relevant passages in Chinese from public sources, and follow the same process to generate outline-enhanced answers. These data will be released to the public after passing the censoring process of data release.

\begin{table*}[t] 
  \caption{The reward model training losses when combining different fine-grained evaluation and fine-grained reward modeling techniques are displayed. Note that our doubly fine-grained RLHF framework provides a unified framework, containing the existing RLHF \cite{christiano2017deep} and fine-grained RLHF approaches \cite{yang2023preference,wu2023fine} as special cases.}
  \label{double_fine_grained_rlhf}
  \begin{center}
  \begin{tabular}{c|ccccc} 
  \toprule
    \multirow{2}{*}{\makecell{Reward model\\granularity}} & \multicolumn{5}{c}{Evaluation Granularity}\\ \cline{2-6}
    & Holistic && Sentence-level && Subclaim-level \\ \midrule %
    Sequence-level & Logloss($\hat{\boldsymbol{R}}_\phi, r$)\cite{christiano2017deep} && $\sum_{i=1}^L$  Logloss($\hat{\boldsymbol{R}}_\phi[i], \boldsymbol{r}_i$)\cite{wu2023fine} && $\sum_{i=1}^L$ MSE($\hat{\boldsymbol{R}}_\phi[i], \textrm{Agg}_j(\boldsymbol{R}_{ij})$) \\
    \midrule
    Token-level & Logloss($\textrm{Agg}_t(\hat{r}_t), r$)\cite{yang2023preference} && $\sum_{i=1}^L$ Logloss($\textrm{Agg}_t(\hat{r}_t)[i], \boldsymbol{r}_i$) && $\sum_{i=1}^L$ MSE($\textrm{Agg}_t(\hat{r}_t)[i], \textrm{Agg}_j(\boldsymbol{R}_{ij})$)\\
    \bottomrule
\end{tabular}
\end{center}
\end{table*}

\section{Factuality-Optimized RAG}
In this section, we propose a novel factuality optimization method to address the aforementioned factuality issue in web-enhanced LFQA. Specifically, we first discuss the difficulty of directly applying the conventional RLHF method to factuality optimization, then develop a novel doubly fine-grained RLHF framework, which characterizes different granularities of automated evaluation and reward modeling, upon which our method is built.

\subsection{Difficulties of Directly Applying RLHF}

In LLM alignment, reinforcement learning with human feedback (RLHF) \cite{christiano2017deep,ouyang2022training} is a widely used technique to reduce undesirable generations, e.g., harmful responses in chat assistant tasks \cite{bai2022training}. Viewing nonfactuality as a certain kind of undesirable behaviors, a natural way to promote factuality in web-enhanced RAG is to utilize RLHF to prevent the generator from producing nonfactual responses. To proceed, we first give a detailed description of RLHF.

Conventionally, RLHF is conducted on manually annotated preference data. For example, given the query $x$ and the retrieved context $z$, the factuality of an answer $\va_T$ (tokenized as $(a_1, \dots, a_T)$) can be annotated as $r  \sim \mathcal{R}(\va_T, x, z)$, where $r \in [0,1]$ reflects the underlying human preference. RLHF trains a \textit{reward model} $\hat{R}$ to estimate the factuality given any query $x$, reference $z$, and answer $\va$, i.e., to learn the human preference function $\mathcal{R}$. Then RL methods such as PPO are applied to optimize the generation model based on the trained reward model $\hat{R}$. The optimization problem can be formulated as 
\begin{align*}
    \max_\theta \sum_{s_1\sim \gD} \sum_{t=1}^{T} \E_{a_t \sim \pi_\theta(\cdot|s_t))} & [\1(t=T)r]  -\ \beta \sD_{\textrm{KL}}(\pi_\theta(\cdot|s_t) || \pi_\textrm{ref}(\cdot|s_t)), \label{rlhf_equation} 
\end{align*}
where $\1$ is the indicator function, $\sD_{\textrm{KL}}$ is KL divergence regularization, and $\beta$ is the regularization strength. In the above formulation, the regularization term is introduced to prevent the generation model from deviating too far from a reference model $\pi_\textrm{ref}$. The reference model is often set as the model after SFT (e.g., our outline-enhanced RAG model as proposed above).

Directly applying the conventional RLHF method to factuality optimization in web-enhanced LFQA will encounter two intrinsic difficulties. First, the manually annotated factuality labels are typically expensive to collect, which involves comparing the factual details between a long answer and its corresponding lengthy reference. Second, as shown in the above equation, the standard RLHF uses the holistic reward, i.e., $\1(t=T)\boldsymbol{R}(\va, x,z)$, which is not zero only for the last token of the whole response. This holistic reward can only provide a sparse signal for the training of the generation model $\pi_\theta$. In web-enhanced LFQA where the answers are usually longer, the sparsity problem due to the use of the holistic reward will be even exaggerated.

\subsection{Doubly Fine-grained RLHF} \label{evaluation}
In light of the above difficulties of conventional RLHF in factuality optimization for web-enhanced RAG, we propose a \textit{doubly fine-grained RLHF} framework to conduct factuality optimization in a fine-grained manner. Our framework is inspired by recent study on fine-grained RLHF \cite{wu2023fine,yang2023preference}. Unlike these previous works that mainly focus on a single dimension, our framework incorporates fine-grained designs of two core steps, i.e., factuality evaluation and reward modeling. 

Before elaborating our framework in details, we first introduce necessary notations and definitions, which enables to characterize multiple rewards for an answer that constitute a denser reward signal $\boldsymbol{R}_\phi$ for the RL process. Specifically, following \cite{wu2023fine}, we first segment the output $\va$ into $L$ text spans $(\va^1, \va^2, \dots, \va^L)$ corresponding to the evaluation granularity (which will be described later) of $\boldsymbol{R}_\phi$, where each segment $\va^j$ ends at the step $T_j$. The dense reward signal $\boldsymbol{R}_\phi$ is an $L$-dimensional vector, whose $j$-th dimension represents the reward $\boldsymbol{R}_\phi(\va|x,z)[j] \in [0,1]$ for each segment $\va^j $ given query $x$ and retrieved context $z$ as the input, which is assigned to the final token in $\va^j$. Especially, when $L=1$, our method degenerates to the standard RLHF with holistic reward.

% Previous studies have further shown that such a technique can improve the quality of automatic evaluation pipeline using LLM such as GPT-4. In this spirit, the main decision factor is the granularity of the factuality evaluation and the corresponding RLHF process. 
\textbf{Fine-grained Evaluation.} Recall that to perform high quality automatic factuality evaluation, recent methods have been proposed to first decompose a long answer into shorter pieces and then evaluate the factuality of each piece with respect to the given reference \cite{kamoi2023wice,lattimer2023fast}. Inspired by these methods, we consider three different levels of granularity in the answer decomposition and automated segment evaluation:

\begin{itemize}
    \item \textit{Holistic}: It is the standard granularity to evaluate the answers \cite{christiano2017deep}.  Each generated answer is associated with a single factuality score $r$. 

    \item \textit{Sentence-level}: As is suggested by previous research on automatic evaluation \cite{kryscinski2019evaluating,lattimer2023fast}, we can segment the answer into sentences,\footnote{We uses pySBD \url{https://github.com/nipunsadvilkar/pySBD}  for segmentation.}  then evaluate each sentence individually. In this case, the evaluation result is denoted as $\boldsymbol{r}_i$ where $i$ is the index for the sentence. 

    \item \textit{Subclaim-level}: Following \cite{kamoi2023wice,chern2023factool,min2023factscore}, we can further decompose each sentence into multiple \textit{subclaims} via an LLM, each containing a single piece of factual information (see Appendix~\ref{app:prompt_decomposition} for the prompts we use). 
    After the decomposition, we evaluate each subclaim individually. Since the decomposition using LLM breaks the association between the subclaim and the original answer, we aggregate the scores of all subclaims into a single score to evaluate the factuality of the sentence. More specifically, assuming there are $j$ subclaims for sentence $i$, then the evaluation score for the sentence is given as $\boldsymbol{r}_i = \textrm{Agg}_j(\boldsymbol{R}_{ij})$, where $\boldsymbol{R}_{ij}$ denotes the factuality score of the subclaim $j$ of sentence $i$, and Agg is the aggregation function (in the form of average, minimum, or maximum).
\end{itemize}

\textbf{Fine-grained Reward Modeling.} Recall that to build a reward model to estimate the factuality of a given answer, standard RLHF methods typically use a sequence-level reward model that produces a single factuality score for each answer. Recently, a token-level reward modeling method has been introduced to provide token-level feedback \cite{yang2023preference}. Enlightened by these methods, we can construct the reward model in two possible levels of granularity.
\begin{itemize}
    \item  \textit{Sequence-level}: A single reward $\hat{\boldsymbol{R}}_\phi(\va|x,z)$ is learned for each sequence, whose actual form depends on the granularity of the evaluation. In this way, the associated reward reflects the factuality of the corresponding sequence, which is then assigned to the last token of each sequence. 
    \item \textit{Token-level}: A reward $\hat{r}(s_t, a_t)$ learned for each token in the sequence. In this way, the reward of the sequence is calculated by aggregating the all token-level rewards i.e., $\hat{\boldsymbol{R}}_\phi(\va|x,z) = \textrm{Agg}_t(\hat{r}(s_t,a_t))$. %where \citet{yang2023preference} shows that the aggregation function $\textrm{Agg}$ taking the average function performs best. 
\end{itemize}
The training loss of each combination of automated evaluation and reward modeling in different levels of granularity is illustrated in Table \ref{double_fine_grained_rlhf}. 
In most cases, the reward $r$ is binary labelled, and the reward model is trained with the Logloss: 
\begin{align*}
\textrm{Logloss}(\hat{\boldsymbol{R}}_\phi, r)=-\sum_{(x,z,\va,r)\in\mathcal{D}}\sum_{i=1}^L \quad \boldsymbol{r}_i\log({\hat{\boldsymbol{R}}_\phi(\va|x,z)[i]})\\
+(1-\boldsymbol{r}_i)\log(1-\hat{\boldsymbol{R}}_\phi(\va|x,z)[i]).
\end{align*}
When the aggregation step in subclaim-level evaluation yields continuous\-/valued reward $r$, we choose to use the MSE loss for reward model training instead 
\begin{align*}
\textrm{MSE}(\hat{\boldsymbol{R}_\phi}, r )=\sum_{(x,z,\va,r)\in\mathcal{D}}\sum_{i=1}^L \quad (\hat{\boldsymbol{R}}_\phi(\va|x,z)[i]-\boldsymbol{r}_i)^2.
\end{align*}

After the reward model $\hat{\boldsymbol{R}}$ is learned, we adopt PPO to optimize the generation model by maximizing the following reward 
\begin{align*}
\hat{r}_t(s_t, a_t)=\sum_{j=1}^{L}\1(t=T_{j})\hat{\boldsymbol{R}}_\phi(\va|x,z)[j]-\beta\log\frac{\pi_{\theta}(a_{t}|s_{t})}{\pi_{\textrm{ref}}(a_{t}|s_{t})}.
\end{align*}
Compared to the conventional RLHF with a single reward for each answer, our formulation has $L$ non-zero rewards corresponding to the segments, which alleviates the sparse feedback signal problem in conventional RLHF. 

Note that our proposed framework unifies the existing fine-grained RLHF works \cite{wu2023fine,yang2023preference} by containing these methods as special cases. Moreover, although our framework is motivated by optimization factuality for web-enhanced RAG, it can also be generalized to other RLHF tasks. 

% \subsection{Factuality Evaluation}

% To evaluate the factuality, we introduce our automatic evaluation of factuality on various granularity levels. For holist-level, we ...

% For sentence and subclaims levels, we uses a two step approaches for evaluation, which first decompose the answers into pieces, and then assess the pieces individually. To decompose the answer into subclaims, we follow \cite{} to first split the answer into sentences, and then use GPT4 to split the answer into subclaims, where similar to existing work, instructions are designed to divide long paragraph or sentences into multiples segments that can be understood independently.

% \paraphrase{We carefully designed a prompt (see Appendix \ref{}) with the aforementioned principles and few-shot examples to decompose of answers into segments using GPT-4. To enhance GPT-4's performance to segment various answers, our few-shot examples include various situations, such as cases that necessitate pronoun resolution, instances with unique formatting, and examples that require special handling of discourse links.\\
% We employ GPT-4 within a few-shot framework to decompose answers in the training dataset, creating a segmented training dataset used to train ChatGLM3 as our own context segmenter. The detailed prompt is provided in Appendix~\ref{ContextSegmenter prompt}.}

\section{Experiment}

In this section, we conduct extensive experiments to validate the effectiveness of our outline-enhanced generation technique and factuality optimization methods.

\subsection{Experimental setup}
\textbf{Datasets.} We conduct experiments on two commonly used datasets for web-enhanced long-form QA. 

\emph{The WebGPT's dataset}. Although the training dataset originally used for WebGPT is not publicly available, the 272 samples released on the WebGPT demo website\footnote{\url{https://openaipublic.blob.core.windows.net/webgpt-answer-viewer/index.html}} can be used as a testbed for performance comparison \cite{liu2023webglm}. In this dataset, each sample consists of a question from the ELI5 dataset \cite{eli5_lfqa}, several Bing retrieved web pages, and extracted references. Note that it is a pure English dataset. 

\emph{The WebCPM's dataset}. \cite{qin-etal-2023-webcpm}. This is a Chinese dataset constructed similarly to the WebGPT dataset. As there is no official train-test split, we randomly split 4,676 samples for training, 426 for validation, and 398 for testing. 

%To test models on Chinese, we conduct further evaluations on the WebCPM dataset \cite{qin-etal-2023-webcpm}. The WebCPM is constructed similarly to the WebGPT dataset. Since there are no official train test split, we randomly split 4,676 samples for training, 426 for validation and 398 samples into the test split. All samples are in Chinese, and each contains question, web page, reference, and human-written answers.

\begin{table*}[t] 
  \caption{Performance comparison of the existing web-enhanced RAGs with our FoRAG.} \label{table_compare}
  \begin{center}
  \begin{tabular}{cccccccccccccc} 
  \toprule
    \multicolumn{1}{c}{\multirow{3}{*}{\makecell{Model}}}  & \multicolumn{11}{c}{\textbf{Answer Evaluation}} \\
    & \multicolumn{5}{c}{\textbf{WebCPM (zh) }} && \multicolumn{5}{c}{\textbf{WebGPT (en)}}\\ \cline{2-12}
     & Cohr. & Help. & Fact/q. & Fact/s. & Avg. Len. && Cohr. & Help. & Fact/q. & Fact/s. & Avg. Len. \\ \midrule % $\downarrow$
    WebGPT 175b    & - & - & - & - & - &&  0.6911  & \underline{0.9154} & \underline{0.8823} & 0.9752 &  209 \\
    WebGPT 13b     & - & - & - & - & - &&  0.5478  & 0.7390 & 0.7977 & 0.9642 &  212 \\
    WebGLM 10B     & - & - & - & - & - &&  0.5919  & 0.8566 & 0.8639 & 0.9688 & 169 \\
    WebCPM 10B     & 0.4899  & 0.6985 & 0.6784 & 0.8916 &  549 && 0.7316  & 0.8566 & 0.8125 & 0.9764 &  330 \\
    FoRAG-C 6B (Ours) & \underline{0.8618}  & \underline{0.7764}  & \underline{0.7739} &\underline{0.9639} &  655 && \underline{0.8603} & 0.8640 & 0.7610 & \underline{0.9804} & 443\\
    FoRAG-L 7B (Ours) & \textbf{0.9121} & \textbf{0.8668} & \textbf{0.8216} & \textbf{0.9727} & 625 && \textbf{0.9889} & \textbf{0.9595} & \textbf{0.8897} & \textbf{0.9894} & 447 \\
    \bottomrule
\end{tabular}
\end{center}
\end{table*}

\begin{table*}[t]
  \caption{Comparison of variants of FoRAG with or without outline-enhanced (Out. Enh.), factuality optimization (Fac. Opt.).} \label{ablation_two_techniques}
  \begin{center}
  \begin{tabular}{cccccccccccccc} 
  \toprule
    \multirow{3}{*}{\makecell{Model}} & \multirow{3}{*}{\makecell{Out.\\Enh.}} & \multirow{3}{*}{\makecell{Fac.\\Opt.}}  & \multicolumn{11}{c}{\textbf{Answer Evaluation}} \\
    & & & \multicolumn{5}{c}{\textbf{WebCPM (zh) }} && \multicolumn{5}{c}{\textbf{WebGPT (en)}}\\ \cline{4-14}
     &  &  & Cohr. & Help. & Fact/q. & Fact/s. & Avg. Len. && Cohr. & Help. & Fact/q. & Fact/s. & Avg. Len. \\ \midrule
    \multirow{4}{*}{FoRAG-C 6B} & \xmark & \xmark & 0.4598  & 0.6332 & 0.7613 & 0.9081 &  583 && 0.4081  & 0.7721 & 0.7868 & 0.9464 &  177\\
    & \xmark & \cmark & 0.4724  & 0.6407 & 0.8065 & 0.9395  & 585  && 0.5184  & 0.7868 & 0.8566 & 0.9763  & 181\\
     & \cmark & \xmark & 0.8643 & \underline{0.7814} & 0.6055 & 0.9197 & 622 && 0.8566 & 0.8529 & 0.5993 & 0.9530 & 417 \\ 
     & \cmark & \cmark  & 0.8618  & 0.7764 & 0.7739 & \underline{0.9639} &  655 && 0.8603 & 0.8640 & 0.7610 & 0.9804 & 443\\
     \midrule
     \multirow{4}{*}{FoRAG-L 7B} & \xmark & \xmark & 0.4296  & 0.6181 & 0.8090 & 0.8875 &  556 && 0.5221  & 0.8676 & 0.8750 & 0.9728 &  186 \\
     & \xmark & \cmark & 0.4447  & 0.6256 & \textbf{0.8618} & 0.9394  & 570  &&  0.5368  & 0.8860 & \textbf{0.8970} & \underline{0.9818}  & 189\\
     & \cmark & \xmark  & \underline{0.9095} & \textbf{0.8668} & 0.6583 & 0.9345 & 613 && \underline{0.9816} & \underline{0.9559} & 0.7978 & 0.9768 & 424 \\
     & \cmark & \cmark & \textbf{0.9121} & \textbf{0.8668} & \underline{0.8216} & \textbf{0.9727} & 625 && \textbf{0.9889} & \textbf{0.9595} & \underline{0.8897} & \textbf{0.9894} & 447 \\
    \bottomrule
\end{tabular}
\end{center}
\end{table*}

\textbf{Compared Methods.} We compare our method with three web-enhanced baseline methods. 

\emph{WebGPT} \cite{nakano2021webgpt} supports interactive web search for long-form QA. It has two versions, namely WebGPT-13B and WebGPT-175B, where the latter one is the currently state-of-the-art performing model for web-enhanced QA. Note that when comparing with WebGPT, we directly use the responses collected from its website.

\emph{WebCPM} \cite{qin-etal-2023-webcpm} is an open source web-enhanced RAG involving interactive web search. It is the first work on Chinese web-enhanced RAG. It is trained on a dataset which contains 5,500 question-answer pairs in Chinese with references. 

% facutality optimized web-enhanced question answering system . FoQA

\emph{WebGLM} \cite{liu2023webglm} is an open source web-enhanced QA system with human performance. It simplifies the interactive web search approach in WebGPT and WebCPM by a two-step retriever and generator framework. It is trained on the WebGLM-QA dataset, which focuses on English only.

%\paraphrase{\textbf{WebGPT} \cite{nakano2021webgpt} is one pioneer- ing work that supports interactive web search for LFQA. The authors first build a web search interface backed up by Microsoft Bing, then recruit annotators to collect information using the interface to answer questions. After that, they fine-tune GPT-3 (Brown et al., 2020) to imitate human behaviors for web search and to organize the collected information into answers. In the experiments, WebGPT shows exceptional ability in LFQA, even surpassing human experts. However, the WebGPT's interface, dataset, and trained models are not publicly available}. Therefore when comparing with WebGPT, we use the retrieved reference and generated responses directly. \textbf{WebCPM} is an open source web-enhanced RAG that involves interactive web search, and it is first work that targets on Chinese web-enhanced RAG. The WebCPM model is trained on a dataset which contains 5,500 question-answer pairs in Chinese with references. \textbf{WebGLM} is an open source web-enhanced QA system with human performance. It simplify the interactive web search approach adopted by the WebGPT and WebCPM by proposing a two-steped retriever and generator framework, and show promising results over existins work. It is trained on the WebGLM-QA dataset, which focuses on English only.

\textbf{Metrics.} We adopt three commonly used metrics for web-enhanced RAG, i.e., \emph{coherence}, \emph{helpfulness}, and \emph{factuality}. As existing works show that GPT4's evaluation is highly consistent to human annotations in both English \cite{G-Eval,Jiang2023TIGERScoreTB,fu2023gptscore,bai2023benchmarking, guo2023close} and Chinese \cite{xu2024face4rag}, we use GPT4 to evaluate these metrics. For the completeness of our study, we also justify the consistency between GPT4 and human annotation in Chinese in ablation study. Following the framework of \cite{G-Eval}, we evaluates the coherence (\textit{Cohr.}) and helpfulness (\textit{Help.}) metrics. We count the scores greater than or equal to 4 as the judging criteria. For evaluation of factuality consistency, we adopt the method in \cite{felm} to achieve fine-grained evaluation. In addition, since the longer answers are more likely to have factuality mistakes, for the fairness of the evaluation, we report the scores at two granularities, i.e., query-level (\textit{Fact/q.}) and sentence-level (\textit{Fact/s.}). The prompts we used for the evaluations are given in Appendix~\ref{app:prompt_evaluation}.

%Note that since the increasing length increase the probabitity to making mistakes, to make a fair comparison, for facutal consistency, we report the sentence-level facutal consistency score as evaluated by GPT4. The prompts we used for evaluation is in Appendix \ref{}. 

% and use GPT4 to evaluate the scores of these metrics. For factuality evaluation, we adopt the evaluation as in \cite{felm}, where they show that GPT4 achieves a high coherence with human annotators in English. T For factuality, we take the fine-grained evaluation by GPT4. \TODO{add on the evaluation metrics. with citations.} The evaluation prompts are provided in Appendix \TODO{\ref{}}. \TODO{explain on the sentence level.}

\textbf{Models and Training Configuration.} 
Our experiments are conducted by fine-tuning on Llama2-7B-chat \cite{touvron2023llama} and ChatGLM2-6B \cite{du2022glm}, which are widely used LLMs for question-answering in English and Chinese respectively. The prompt templates at fine-tuning and inference stages are given in Appendix \ref{app:prompt_geneartion}. The maximum context length is set to 4096 for Llama2-7B-chat and 8192 for ChatGLM2-6B. Both models are fine-tuned on 8 A100 GPUs for 5 epochs with a initial learning rate of 1e-5 and a cosine learning rate scheduler. Following the configuration of WebCPM \cite{qin-etal-2023-webcpm}, we adopt beam search for each inference on a single A100 GPU with the \emph{num\_beams} parameter set to 3. We use our outline-enhanced dataset to conduct supervised fine-tuning (SFT) \cite{ouyang2022training}, and our multi-granularity evaluation data to conduct corresponding factuality optimization. In order to decrease noise in the RLHF step, we normalize the reward. Specifically, for each prompt $(x,z)$, we generate a response $\va_T$ using the SFT model, and estimates its reward score  $\hat{R}(\va_T,x,z)$ using the learned reward model. For any model generated answer $\va'_{T'}$, we take $\tilde{R}(\va'_{T'}, x, z) = \hat{R}(\va'_{T'}, x, z)-\hat{R}(\va_T, x, z)$ as the estimated reward, and the same technique is applied to sentence-level and subclaim level factuality evaluations.

% perform preliminary test on four backbone models, i.e. Mistral-7B \cite{}, Llama2-7B-chat \cite{}, Qwen-7B-chat and Chatglm2-6B, where the later two are enhanced specifically for question answering in Chinese. On both open-sourced dataset and our collected dataset, it seems

% Llama 2 is a collection of pretrained and fine-tuned generative text models ranging in scale from 7 billion to 70 billion parameters. This is the repository for the 7B fine-tuned model, optimized for dialogue use cases and converted for the Hugging Face Transformers format. Links to other models can be found in the index at the bottom.

% Model Details
% Note: Use of this model is governed by the Meta license. In order to download the model weights and tokenizer, please visit the website and accept our License before requesting access here.

% Meta developed and publicly released the Llama 2 family of large language models (LLMs), a collection of pretrained and fine-tuned generative text models ranging in scale from 7 billion to 70 billion parameters. Our fine-tuned LLMs, called Llama-2-Chat, are optimized for dialogue use cases. Llama-2-Chat models outperform open-source chat models on most benchmarks we tested, and in our human evaluations for helpfulness and safety, are on par with some popular closed-source models like ChatGPT and PaLM.

\subsection{Main results}

The main empirical results of our method trained on Llama2-7B-chat (FoRAG-L 7B) and ChatGLM2-6B (FoRAG-C 6B) are depicted in Table \ref{table_compare} and Table \ref{ablation_two_techniques}. We here mainly report the results of FoRAG-L 7B, which attains the best performance among all possible combinations of the granularities of evaluation and reward model. A detailed performance comparison of different granularity combinations will be given in Section \ref{ablation}.

% ... +... 的实验结果，因为我们实验中发现这个配置在。。。中的效果最好。关于这些配置之间的具体性能比较，请见后面的ablation study部分。

\textbf{Overall Performance}. In Table \ref{table_compare}, we compare the overall performance of FoRAG-L 7B and FoRAG-C 6B with all existing methods on both datasets. Note that among the examined baselines,  only WebCPM 10B can answer in Chinese, since WebGPT does not release model weights or answers in Chinese, and WebGLM has a well-known issue of being unable to answer in Chinese\footnote{Please refer to the discussion on the issue at WebGLM's official codebase \url{https://github.com/THUDM/WebGLM/issues/7}.}. From the results, we observe that on both English and Chinese datasets, FoRAG-C 6B surpasses all baselines on five out of six metrics, and FoRAG-L 7B performs the best on all metrics. Notably, FoRAG-L 7B substantially outperforms WebGPT 175B that contains 24 times more parameters, showing superiority of our method in bilingual web-enhanced RAG tasks. 

%all existing methods, including the WebGPT 175B despite it contains 24 times more parameters. Our FoRAG-C 6B also surpass all existing methods on five out of six metrics. 

\textbf{Evaluation of Outline-Enhanced Generator}. We evaluate the effectiveness of outline-enhanced generator as a core design in dataset collection and RAG model design by showing that, without such technique, the reduced method will deteriorate severely. Specifically, for the reduced variant, we train the backbone models on a merged dataset from WebCPM and WebGLM-QA, which contains 4.7k samples in Chinese and 44k samples in English. Note that here we use the demonstration answers as provided in the original datasets, i.e., the answers in WebCPM are human written and answers in WebGLM-QA are GPT4 generated, which can be considered in high quality. 

In Table~\ref{ablation_two_techniques}, we compare the performance of our proposed method with the reduced variant. The results show that applying our technique of outline-enhanced generator significantly boosts the performance in terms of coherence and helpfulness on both datasets. As for factuality, the sentence-level measurements of our methods are a little bit higher or comparable with the counterpart models without outline-enhanced techniques. In addition, applying our technique increases the length of model generations (see Appendix \ref{example_our_model_baseline} for examples of generations).

%In Table \ref{ablation_two_techniques}, we compare the performance of training our method with or without or outline-enhanced generator techniques. For the without outline-enhanced group, we trained the backbone models using all available open-source datasets, i.e. the WebCPM and WebGLM-QA dataset together, which contains 4.7k data in Chinese, and 44k data in English, respectively. The demonstration answer in these two dataset are human annotated and GPT4 generated, and therefore can be considered as high-quality. However, neither apply the outline-enhanced techniques. From Table \ref{ablation_two_techniques}, it is clear that applying our outline-enhanced techniques significantly boost the performance of both models on the coherence and helpfulness metrics on both datasets. It also guide the model to generate longer answers, which has been shown to be more preferred by human annotators \cite{}.  The sentence-level facutal consistency metrics are a little bit higher or comparable with the models trained on existing datasets, which suggests that directly improving the quality of generated answer may not improve the factuality of the generation model. 

\textbf{Evaluation of Factuality Optimization}. We then evaluate the effectiveness of the factuality optimization technique by comparing our method with the counterpart method without such a mechanism. As presented in Table~\ref{ablation_two_techniques}, adding factuality optimization technique significantly raises the factuality consistency score in both query and sequence levels, without affecting the other two metrics or the generation length. The above results justify the introduction of factuality optimization technique to our proposed method. 

%We then apply the subclaim-level RLHF to the models trained with our outline-enhanced or open source datasets models, which performs best in our ablation study on factuality optimization. Applying this factuality-optimization significantly improves the models' factuality, on all the sets, with little affect on other metrics, such as the coherence, helpfulness and generation length. 

\begin{table*}[t] 
  \caption{The comparison of performance using various factuality optimization techniques on our FoRAG-L 7B. The ``-" indicates no extra factuality optimization is performed, i.e., the SFT model.}
  \label{factuality_optimization}
  \begin{center}
  \begin{tabular}{ccccccccccccc} 
  \toprule
    \multicolumn{2}{c}{\multirow{3}{*}{\makecell{Factuality\\Optimization}}}  & \multicolumn{11}{c}{\textbf{Answer Evaluation}} \\
    & & \multicolumn{5}{c}{\textbf{WebCPM (zh) }} && \multicolumn{5}{c}{\textbf{WebGPT (en)}}\\ \cline{3-13}
    & & Cohr. & Help. & Fact/q. & Fact/s. & Avg. Len. && Cohr. & Help. & Fact/q. & Fact/s. & Avg. Len. \\ \midrule % $\downarrow$
    \multicolumn{2}{c}{-} & 0.9095 & \textbf{0.8668} & 0.6583 & 0.9345 & 613 && 0.9816 & 0.9559 & 0.7978 & 0.9768 & 424\\ 
    \midrule
     \multicolumn{2}{c}{Unlikelihood} & 0.9070 & 0.8618 & 0.7286 & 0.9477 & 591 && 0.9816 & 0.9522 & 0.8419 & 0.9794 & 442 \\
     \multicolumn{2}{c}{MLE w. Filtering} & \textbf{0.9171} & 0.8568 & 0.6783 & 0.9331 & 592 && 0.9852 & 0.9522 & 0.7831 & 0.9754 & 423 \\
    \midrule
    \multirow{6}{*}{RLHF} & Holistic + Token & 0.9020 & 0.8543 & 0.7236 & 0.9414 & 608 && 0.9816 & 0.9485 & 0.8382 & 0.9768 & 444 \\
    & Sentence + Token & 0.9095 & 0.8593 & 0.7814 & 0.9628 & 610 && 0.9816 & 0.9559 & 0.8603 & 0.9836 & 446 \\
    & Subclaim + Token & {0.9121} & 0.8593 & 0.7864 & 0.9658 & 616 && 0.9852 & 0.9559 & 0.8713 & 0.9851 & 446\\
    & Holistic & 0.9146 &  0.8618 & 0.7563 & 0.9526 & 622 && 0.9852 & 0.9559 & 0.8493 & 0.9797 & 448 \\
    & Sentence & 0.9095 & 0.8593 & 0.8065 & 0.9704 & 612 && \textbf{0.9889} & \textbf{0.9595} & 0.8787 & 0.9866 & 447 \\
    & Subclaim & 0.9121 & \textbf{0.8668} & \textbf{0.8216} & \textbf{0.9727} & 625 && \textbf{0.9889} & \textbf{0.9595} & \textbf{0.8897} & \textbf{0.9894} & 447\\
    \bottomrule
\end{tabular}
\end{center}
\end{table*}

\subsection{Comparison of Various Factuality Optimization Granularities} \label{ablation}
In this subsection, we compare the performance of various implementations of our method on different evaluation and reward model granularities (as described in Table \ref{double_fine_grained_rlhf}), with the following two commonly used alignment methods as baselines:

\emph{MLE with Filtering (filter.)} \cite{solaiman2021process}: This method applies a filter to drop the samples with factual inconsistency errors and preserve the factually consistent ones. Then it follows the standard SFT procedure, i.e., fine-tuning the model by optimizing the maximum likelihood estimation loss on the positive samples. 

\emph{Unlikelihood} \cite{welleck2019neural}: This method fine-tunes the model by maximizing the likelihood of positive (i.e., factually consistent) samples and minimizing the likelihood of negative (i.e., factually inconsistent) samples simultaneously. %In our experiments, we treat samples that are factual consistent as positive samples, and samples with factual inconsistency errors as negative samples. 

For a fair comparison, all the methods are fine-tuned from the same model, i.e., Llama2 after SFT on our outline-enhanced dataset. The empirical results are presented in Table \ref{factuality_optimization}. From the results, we observe that our proposed method attains better factual consistency than the baselines, regardless of the granularity of evaluation or reward modeling. In addition, among all the granularities of evaluation, subclaim-level evaluation performs the best. We also notice that token-level reward modeling performs worse than the conventional segment-level reward modeling, presumably because the length of our datasets may make token-level modeling over-fit. 

%Our factuality optimization techniques have little effects on the other metrics.  For the other two baselines, MLE w. Filtering seems have no improvement on the performance which coincide the previous results that improving the dataset quality does not improve the factuality of resulting models. The unlikelihook methods improves the factuality a little but is outperformed by most of the RLHF-based factuality optimization techniques. 

\subsection{Ablation study}
We now conduct ablation study to justify the rational of some certain design choices in our proposed method. 

%In this section, we study the major improvements and strategies of our method, including the effect of our outline-enhanced generation technique, the importance of data balance in our bilingual generation task and some other minor topics.

\textbf{Effectiveness of the Outline-Expansion Two-Step Answer Generation.}
To illustrate the impact of our outline-enhanced generation technique, we train two baseline models that generate answers directly based on our dataset, which lack the outline stage, referred to FoRAG-C 6B w/o outline and  FoRAG-L 7B w/o outline in Table \ref{outline_effectiveness}. The outcomes clearly show that our outline-enhanced generation approach significantly augments the model's capabilities by enhancing the coherence and helpfulness of the answers generated, with a particularly notable improvement observed in the Chinese language task.

\begin{table}[t]
\centering
\caption{Ablation study on outline-enhanced generation.}
\label{outline_effectiveness}
% \resizebox{\columnwidth}{!}{%
\begin{tabular}{@{}ccccc@{}}
\toprule
 & \multicolumn{2}{c}{\textbf{WebCPM (zh)}} & \multicolumn{2}{c}{\textbf{WebGPT (en)}} \\ \cmidrule(l){2-5} 
Method & Cohr. & Help. &  Cohr. & Help. \\ \midrule
FoRAG-C 6B w/o outline & 0.8467 & 0.7613 & 0.8492 & 0.8529 \\
FoRAG-C 6B & 0.8643 & 0.7814 & 0.8566 & 0.8529 \\ 
\midrule
FoRAG-L 7B w/o outline & 0.8543 & 0.8593  & 0.9779 & 0.9522 \\
FoRAG-L 7B & \textbf{0.9095} & \textbf{0.8668} & \textbf{0.9816} & \textbf{0.9559} \\ 
\bottomrule
\end{tabular}%
% }
\end{table}

\begin{table}[t]
\centering
\caption{Alignment between GPT4-aided automated and human evaluations.}
\label{tab:ablation_gpt4_eval_quality}
\begin{tabular}{@{}ccccc@{}}
\toprule
\textbf{Metrics} & \textbf{Cohr.} & \textbf{Help.} & \textbf{Fact/q.} & \textbf{Fact/s.} \\ \midrule
GPT4 vs Human & 91.5\% & 83.0\% & 77.0\% & 95.5\% \\
Human vs Human & - & - & 69.5\% & 93.1\% \\
\bottomrule
\end{tabular}%
\end{table}

\textbf{On GPT4 Evaluation Quality.}
To evaluate how well GPT4 correlates with human judgment in Chinese, we recruit 10 native Chinese\-/speaking annotators.
Their task is to manually review coherence, helpfulness, and both query-level and sentence-level factuality on the Chinese generated results.
A subset of 200 examples is selected, and we conduct two rounds of human evaluation on it. In each round, each sample is randomly assigned to one annotator.
% For over 90\% of samples, the evaluation results of two rounds are the same showing the reliability of human ratings.
% The agreement between these two rounds is high, with over 90\% consistency.
% We then calculate the Pearson correlation\footnote{Given that our selected metrics are binary, it is expected that the Pearson, Spearman, and Kendall correlation would be identical.} between GPT4 and average human ratings on these four metrics, as shown in Table~\ref{tab:ablation_gpt4_eval_quality}.
We report the agreement rate (the ratio of overlap) between two-round human labels and GPT4\'s judgements in Table~\ref{tab:ablation_gpt4_eval_quality}.
The results confirm a robust correlation between GPT4 and human ratings on Chinese QA evaluation. Except on the query-level factuality, human suffers from comparing two lengthy texts, a conclusion that is consistent with \cite{wu2023fine}.

\begin{figure}[tb] 
  \centering
  \includegraphics[width=\linewidth]{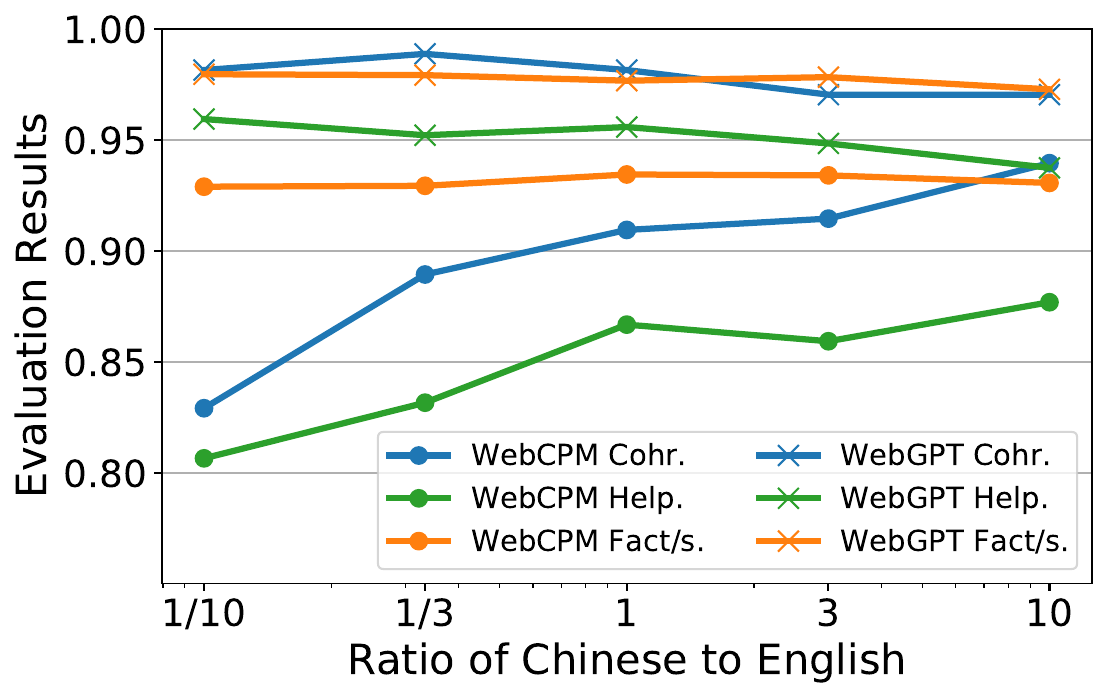}
  \caption{Evaluation results in terms of various metrics of different models fine-tuned from Llama2-7B. We vary the ratio of the Chinese samples to the English samples in the training dataset.} \label{effect_of_imbalance_dataset}
\end{figure}

\textbf{Effects of Imbalance Dataset.}
To evaluate how the imbalance of the two languages in the dataset affects the training effect on the resulting bilingual LLMs, we perform further ablation study on the level of imbalance. We fix the number of training samples to be 40k. Then we tune the ratio of Chinese to English on five level, ranging from 1:10, 1:3, 1:1 to 3:1, 10:1. We then randomly sample the corresponding amount of samples from our dataset and train the models based on Llama2-7B using SFT. The evaluation results, as depicted in Figure \ref{effect_of_imbalance_dataset} show that with a increasing amount of data, the model's performance on the corresponding language increases on both coherency and helpfulness metrics. Meanwhile, the factuality metric is not affected by this ratio. Note that the performance of Llama2-7B is more sensitive to the number of training samples in Chinese. This may be because the pre-training of Llama2-7B contains more corpus in English, and therefore a few examples is enough to adopt to the new task.

% \paragraph{Compare with exsiting methods}

% We compare the generation helpfulness of our method with various baselines. The resutls as shown in Figure \ref{win_rate}, show that the usage of draft before generation significantly improves the preference level of the response of our methods. 

\subsection{Evaluation of Training Efficiency}

We finally evaluate the training efficiency of our proposed method. In the following, we will examine the additionial computation cost of the two new modules of FoRAG, i.e., outline-enhanced generation and doubly fine-grained RLHF, respectively. 

The first step, i.e., outline-enhanced generation, has almost negligible effect on training time. During inference, it requires roughly 10\% more tokens to be generated, and the extra time consumed at inference stage is roughly proportionally to this increase in tokens generated. Note that this extra inference time can be eliminated using context distillation techniques\citep{welleck2019neural,askell2021general}.

The second step, i.e., doubly fine-grained RLHF, has no impact on inference time. To evaluate the additional computational expense during training, we consider a naive implementation of FoRAG that sequentially evaluates the reward for each sentence. In Table \ref{rlhf_efficiency}, we compare the training time of FoRAG and the counterpart method with holistic RLHF. The results show the best performed version, the subclaim version of the doubly fine-grained RLHF framework, takes about 67.7\% more time than standard RLHF. Note that the additional computational cost can be further reduced via implementation with a multi-head reward layer and carefully designed attention mask can use one forward pass to calculate the reward for all sentences, which will make the extra computational cost insignificant.

In summary, FoRAG outperforms the baseline method with reasonable additional computational cost. 

\begin{table}[t]
\centering
\caption{Evaluation on training efficiency of doubly fine-grained RLHF. We report the time consumed (in hours) for training reward models (RM) and RLHF on different granularities. We also report the extra time cost in percentage compared with the holistic RLHF method.}
\label{rlhf_efficiency}
  \begin{tabular}{ccccc} 
  \toprule
    & RM & RLHF & Total & Percentage\\ \midrule
    Holistic & 1.1 & 32.0 & 33.1 & - \\
    Sentence & 4.6 & 45.6 & 50.2 & +51.7\%\\
    Subclaim & 5.1 & 50.4 & 55.5 & +67.7\%\\
    Holistic + Token & 1.3 & 32.8 & 34.1 & +3.0\% \\
    Sentence + Token & 5.3 & 46.4 & 51.7 & +56.2\%\\
    Subclaim + Token & 6.0 & 52.0 & 58.0 & +75.2\%\\
    \bottomrule
\end{tabular}

\end{table}

\section{Conclusion} 
In this paper, we propose a novel answer generation method FoRAG for web-enhanced LFQA to tackle the factuality issue and lack of clear logical structure in existing methods. To this end, we first devise an outline-enhanced generator to fulfill clear logic in long-form answers and accordingly construct two datasets. Then we propose to optimize factuality in a carefully designed doubly fine-grained RLHF framework. Our developed framework contains automatic evaluation and reward modeling in different levels of granularity, and compasses traditional fine-grain RLHF methods as special cases. Empirically, FoRAG achieves state-of-the-art performance in terms of coherence, helpfulness, and factuality on both English and Chinese benchmarks. Notably, applying FoRAG to Llama2-7B-chat, we derive FoRAG-L-7B, which outperforms WebGPT-175B with only 1/24 in the number of parameters of WebGPT-175B. 
% \clearpage

% \begin{table*}
%   \caption{Some Typical Commands}
%   \label{tab:commands}
%   \begin{tabular}{ccl}
%     \toprule
%     Command &A Number & Comments\\
%     \midrule
%     \texttt{{\char'134}author} & 100& Author \\
%     \texttt{{\char'134}table}& 300 & For tables\\
%     \texttt{{\char'134}table*}& 400& For wider tables\\
%     \bottomrule
%   \end{tabular}
% \end{table*}

%%
%% The next two lines define the bibliography style to be used, and
%% the bibliography file.
\bibliographystyle{ACM-Reference-Format}
\bibliography{sample-base}

%%
%% If your work has an appendix, this is the place to put it.

% \clearpage

\onecolumn

\appendix

\begin{CJK*}{UTF8}{gbsn}
% \section{More Details in Implementing PPO} \label{ppo}

% In order to normalize the reward to decrease noise, in our training practice, an outcome $y_{0i}$ produced by the base model is provided together with each prompt $x_{i}$. Given the prompt $x_{i}$ and a generated response $y_{i}$, let us denote the score produced by the reward model by $R(x_{i}, y_{i})$. While scoring a generated response $y_{i}$, we take $\tilde{R}(x_{i}, y_{i})=R(x_{i}, y_{i})-R(x_{i}, y_{0i})$ as the final reward, and the generation optimization pipeline is similar to sentence-level. 

\section{Prompts For Generation}
\label{app:prompt_geneartion}

The following is the prompt template used to invoke GPT4 to generate outline-enhanced answers.%, and the same prompt is used for training and testing in all experiments. 

% Chinese Retrieval
% \begin{longtable}{|p{\textwidth}|}
%     \caption{Prompt for Chinese Retrieval}
%     \label{retrieval_prompts_zh}\\
%     \hline
% 你的任务是从\lbrack 文档\rbrack 中抽取与\lbrack 问题\rbrack 相关且能够回答\lbrack 问题\rbrack 的内容, 请直接返回原文的片段，不要进行总结或改写。如果认为文档不能回答问题，则返回“无法回答”。\\
% 示例：\\
% \lbrack 问题\rbrack ：杭州公积金提取额度\\
% \lbrack 文档\rbrack ：2023年1月1日起，杭州无房租赁如何提取住房公积金？具体如何操作？一起来看。\\
% 今年起，无房租赁提取住房公积金限额具体标准：杭州市区（含萧山区、余杭区、临平区、富阳区、临安区）为1500元/月，桐庐县为1050元/月，淳安县为750元/月，建德市为600元/月。\\
% 温馨提示：\\
% 缴存人家庭租住公共租赁住房的，按照实际租金支出提取本人及配偶住房公积金账户余额。\\
% \lbrack 抽取结果\rbrack：无房租赁提取住房公积金限额具体标准：杭州市区（含萧山区、余杭区、临平区、富阳区、临安区）为1500元/月，桐庐县为1050元/月，淳安县为750元/月，建德市为600元/月。\\
% 请根据以上示例进行回答。\\
% \lbrack 问题\rbrack ：\textbf{\{query\}}\\
% \lbrack 文档\rbrack ：\textbf{\{paragraph\}}\\
% \lbrack 抽取结果\rbrack：\\
% \hline
% \end{longtable}

% English Generation
\begin{longtable}{|p{\textwidth}|} 
    % \caption{Prompt for English Generation} 
    % \label{generation_prompts_en}\\
    \hline\#\#\#Task\#\#\#\\
Answer the question based on the materials provided.\\
\\
\#\#\#Requirements\#\#\#\\
Step One: Develop an answer outline based on the question and materials.\\
1. Choose a suitable organizational pattern for the answer structure, such as general-specific-general, progressive, comparative, cause-effect, parallel, chronological, among others.\\
2. Enumerate the essential points that need to be included in the outline, aligned with the the chosen structure.\\
3. The relationship between key points can be parallel, contrastive, progressive, etc., but should not be repetitive or inclusive.\\
4. Formulate a clear and concise outline that includes at least 1 but no more than 5 key points.\\
5. Each main point should reference only one specific part of the provided materials and must include the material's number within the outline.\\
\\
Step Two: Answer the question based on the materials and outline.\\
1. Utilize the outline as a blueprint to develop a comprehensive and informative answer.\\
2. Write the answer using formatting tools such as numbered lists, bullet points, subheadings, LaTeX formulas, etc., where appropriate.\\
3. Refrain from using basic sequential connectors like ``firstly," ``secondly," or ``furthermore," in the answer.\\
4. Avoid redundancy and repetition of content within the answer.\\
5. Do not cite the number of the materials in the answer.\\
6. Adhere strictly to the information contained within the provided materials, without adding any information that is not included in the materials.\\
\\
\#\#\#Format\#\#\#\\
\lbrack Structure\rbrack :\\
<The organizational structure of the answer>\\
\lbrack Outline\rbrack :\\
<Bullet points outlining the answer>\\
\lbrack Answer\rbrack :\\
<Answer the question based on the materials and outline>\\
\\
Here are two examples of input and output:\\
\\
\#\#\#Input Example 1\#\#\#\\
\#\#\#Question\#\#\#\\
What are the main factors contributing to the increasing trend of urban population growth?\\
\\
\#\#\#Materials\#\#\#\\
\lbrack 1\rbrack The improvement of healthcare in cities has led to lower mortality rates. With access to state-of-the-art hospitals, specialized clinics, and a higher concentration of medical professionals, city residents benefit from quicker and more effective response to health needs. The advancements in healthcare infrastructure, such as the implementation of digital health records and telemedicine, have further streamlined patient care, making it more accessible and efficient.\\
\lbrack 2\rbrack Urban areas are widely recognized for better educational institutions, from primary schools to universities, that promise a more comprehensive and diverse learning environment. These institutions are often equipped with advanced resources, experienced educators, and a broader range of programs and extracurricular activities, which are highly attractive to families who prioritize their children's education.\\
\lbrack 3\rbrack Cities are the epicenters of industry, commerce, and services, offering a myriad of employment options that are scarce or nonexistent in rural settings. The diversity of the urban job market, with its potential for higher wages, career advancement, and the chance to engage in more specialized work, draws individuals in pursuit of a better livelihood. Rural to urban migration is driven by the search for better employment opportunities.\\
\\
\#\#\#Output Example 1\#\#\#\\
\lbrack Structure\rbrack :\\
Parallel\\
\lbrack Outline\rbrack :\\
1. Advancements in healthcare (based on \lbrack 1\rbrack )\\
2. Enhanced educational opportunities (based on \lbrack 2\rbrack )\\
3. Employment prospects (based on \lbrack 3\rbrack )\\
\lbrack Answer\rbrack :\\
Urban population growth is driven by several compelling factors that make living in cities increasingly attractive.\\
One of the most significant contributors to this trend is the comprehensive improvement in healthcare that cities offer. Residents benefit from lower mortality rates due to access to superior medical facilities including state-of-the-art hospitals and specialized clinics. Furthermore, the urban healthcare infrastructure has been enhanced by technological advancements, such as the implementation of digital health records and the growing availability of telemedicine services.\\
In addition to healthcare, the quality of education available in cities acts as a strong magnet for families and individuals. Cities typically boast:\\
  - Better educational institutions, which provide a more comprehensive and diverse learning environment.\\
  - Access to experienced educators and advanced educational resources that are not as readily available in rural areas.\\
  - A broad range of extracurricular activities that are highly attractive for those prioritizing education for their children.\\
  Finally, the wide array of employment options available in urban areas is a another driver of population growth. The concentration of industry, commerce, and services in cities creates a diverse job market with the potential for higher wages, career advancement, and specialized work. This draws people from rural areas in search of better employment prospects, thereby increasing the urban population.\\
\\
\#\#\#Input Example 2\#\#\#\\
\#\#\#Question\#\#\#\\
What makes smartphones an essential tool for modern life?\\
\\
\#\#\#Materials\#\#\#\\
\lbrack 1\rbrack Smartphones provide a convenient means of communication, allowing people to stay connected with friends and family through calls, texts, and various messaging apps.\\
\lbrack 2\rbrack They offer access to a wide array of information and entertainment options, such as browsing the internet, streaming videos, and playing games.\\
\lbrack 3\rbrack Smartphones have built-in features that replace the need for other devices, like a camera, GPS, and notebook, making them a multifunctional tool that fits in your pocket.\\
\\
\#\#\#Output Example 2\#\#\#\\
\lbrack Structure\rbrack :\\
General-Specific-General\\
\lbrack Outline\rbrack :\\
1. Communication hub (based on \lbrack 1\rbrack )\\
2. Multifunctional device (based on \lbrack 3\rbrack )\\
\lbrack Answer\rbrack :\\
Smartphones have become indispensable in contemporary life primarily because they serve as a central hub for communication and a versatile multifunctional device.\\
\#\#\# Central Hub for Communication\\
The ability to instantly connect with others via calls, texts, and various messaging applications is one of the most valued features of smartphones. This constant connectivity has revolutionized the way we maintain personal and professional relationships.\\
\#\#\# Versatile Multifunctional Device\\
Smartphones eliminate the need to carry multiple gadgets by incorporating essential tools like a high-quality camera, a reliable GPS navigation system, and the functionality of a notebook, all within a single, portable device.\\
In essence, the critical role of smartphones in modern life can be attributed to their unparalleled capacity to keep us connected to the world around us and their ability to perform multiple functions that simplify our daily routines.\\
\\
Below is the question you need to answer and the corresponding materials. Please output according to the above requirements, format, and examples:\\
\\
\#\#\#Question\#\#\#\\
\textbf{\{query\}}\\
\\
\#\#\#Materials\#\#\#\\
\textbf{\{doc\}}\\
\hline
\end{longtable}

% Chinese Generation
\begin{longtable}{|p{\textwidth}|} 
    \caption{Prompt for Chinese Generation} \label{retrieval_prompts_ch}\\
    \hline
\#\#\#任务\#\#\#\\
根据资料回答问题。\\
\\
\#\#\#要求\#\#\#\\
第一步：根据问题和资料生成回答提纲。\\
1. 决定回答结构，从总分总、递进、对比、因果、并列、时序等结构中选择合适的来组织回答。\\
2. 根据回答结构，在提纲中要完整地列出答案中需要包括的要点。\\
3. 要点之间可以是并列、对照、递进等关系，不可以是重复或者包含关系。\\
4. 要点要保持精炼，至少有1点，不能多于5点，\\
5. 每个要点仅可参考1段资料，并在提纲中标注资料编号。\\
\\
第二步：根据资料和提纲对问题进行回答。\\
1. 回答要以提纲为蓝本，对问题进行详细的回答。\\
2. 回答中可以采用编号或项目列表、小标题、latex公式等格式。\\
3. 回答中减少使用“首先”、“其次”、“再者”等简单的连接词。\\
4. 回答中不要生成重复内容。\\
5. 回答中不要标注资料来源。\\
6. 回答应当严格依据资料，不采用不在资料中的内容。\\
\\
\#\#\#格式\#\#\#\\
【结构】：\\
<回答的组织结构>\\
【提纲】：\\
<分点介绍回答思路>\\
【回答】：\\
<根据资料和提纲回答问题>\\
\\
下面是1个示例输入和2个满足要求的示例输出：\\
\\
\#\#\#示例输入\#\#\#\\
\#\#\#问题\#\#\#\\
2023年西安房贷利率最新消息\\
\\
\#\#\#资料\#\#\#\\
\lbrack 1 \rbrack 一、西安商业贷款固定利率\\
1年以内（含）——4.35\%\\
5年(含)以下——4.75\%\\
5年以上——4.9\%\\
贷款市场报价利率LPR：目前1年期LPR为3.45\%，5年期LPR为4.2\%\\
首套住房商业性个人住房贷款利率下限为不低于相应期限LPR减20个基点。\\
二套住房商业性个人住房贷款利率下限为不低于相应期限贷款市场报价LPR利率加20个基点。\\
二、西安公积金贷款利率\\
5年(含)以下——2.6\%\\
5年以上——3.1\%\\
\lbrack 2 \rbrack 目前西安主流银行的首套房利率集中在4\%左右，二套房利率差基本维持在4.9\%。\\
\lbrack 3 \rbrack 首先，虽然LPR在7月没有变动，但西安首套房贷款利率已经低至4\%，并且低于2009年房贷利率打七折后的4.156\%！\\
这点也恰恰和楼市走访到的信息不谋而合，据了解，西安目前多数银行首套房贷利率主要集中于4\%，二套房贷利率基本在4.9\%左右。\\
\lbrack 4 \rbrack 7月20日，中国人民银行授权全国银行间同业拆借中心公布了最新一期贷款市场报价利率（LPR）：1年期LPR为3.55\%，5年期以上LPR为4.20\%，均与上个月持平。但西安房贷利率较上月小幅下行，西安多家银行首套房贷利率从4.1\%降至4\%。\\
今年6月，LPR时隔10个月迎来下调，1年期和5年期以上LPR均跟随政策利率下调10个基点。记者了解到，从6月下旬开始，西安各大银行相继落实首套房贷利率政策动态调整机制，进行利率调整。目前，西安地区六大行及招商银行、宁夏银行、北京银行、西安银行等十余家银行首套房贷利率降至4\%，较LPR下浮0.2\%。二套房贷利率同步下行，跌破5\%大关，维持在4.9\%。\\
\\
\#\#\#示例输出1\#\#\#\\
【结构】：\\
总分总\\
【提纲】：\\
1. 首套房商业贷款利率(仅使用\lbrack 2 \rbrack 回答)\\
2. 二套房商业贷款利率 (仅使用\lbrack 2 \rbrack 回答)\\
3. 公积金贷款利率（仅使用\lbrack 1 \rbrack 回答）\\
【回答】：\\
西安的房贷市场在2023年呈现了一定的调整趋势，特别是在商业贷款和公积金贷款利率方面。针对不同的贷款年限，首套和二套住房的贷款利率有所不同。下面是具体的利率情况和银行对房贷利率的调整概述。\\
\#\#\# 首套房商业贷款利率\\
首套住房商业贷款利率集中在4\%左右，较LPR下浮0.2\%。\\
\#\#\# 二套房商业贷款利率\\
二套房商业贷款利率基本维持在4.9\%左右。\\
\#\#\# 公积金贷款利率\\
公积金贷款利率5年(含)以下为2.6\%，5年以上为3.1\%。\\
西安的房贷市场在2023年经历了积极的调整，尤其是首套房贷款利率的下降，为购房者带来了更为优惠的借款条件。与此同时，公积金贷款的低利率也为有资格的借款人提供了额外的经济支持。这些调整对于稳定房地产市场、支持合理住房需求有着重要的作用。\\
\\
\#\#\#示例输出2\#\#\#\\
【结构】：\\
并列\\
【提纲】：\\
1. 西安商业贷款固定利率（仅使用\lbrack 1 \rbrack 回答）\\
2. 西安公积金贷款利率（仅使用\lbrack 1 \rbrack 回答）\\
3. 首套与二套房贷款利率最新变化（仅使用\lbrack 4 \rbrack 回答）\\
【回答】：\\
2023年西安的房贷利率最新消息如下：\\
商业贷款固定利率：\\
- 1年以内（含）的利率为4.35\%\\
- 5年（含）以下的利率为4.75\%\\
- 5年以上的利率为4.9\%\\
公积金贷款利率：\\
- 5年（含）以下为2.6\%\\
- 5年以上为3.1\%\\
据报道，7月20日，虽然LPR没有变动，但西安首套房贷款利率已经低至4\%，低于2009年房贷利率打七折后的4.156\%，二套房贷利率同步下行，维持在4.9\%。在6月LPR下调后，西安各大银行相继落实首套房贷利率政策动态调整机制，进行利率调整。\\
\\
下面是你要回答的问题和相应的资料，请根据上述要求、格式和示例进行输出：\\
\\
\#\#\#问题\#\#\#\\
\textbf{\{query\}}\\
\\
\#\#\#资料\#\#\#\\
\textbf{\{doc\}}\\
\hline
\end{longtable}

\clearpage

\section{Prompts for Evaluation}
\label{app:prompt_evaluation}

As mentioned, the followings are the prompts we used to invoke GPT4 to evalute the coherence, helpfulness, and factuality.

% English Coherence
\begin{longtable}{|p{\textwidth}|}
    \caption{Prompt for English Coherence Evaluation}
    \label{coherence_prompts_en}\\
    \hline
You will be given one answer written for a question.\\
Your task is to rate the answer on one metric.\\
Please make sure you read and understand these instructions carefully. Please keep this document open while reviewing, and refer to it as needed.\\
\\
Evaluation Criteria:\\
Coherence (1-5) - the collective quality of all sentences. The answer should have no datelines, system-internal formatting, capitalization errors or obviously ungrammatical sentences (e.g., fragments, missing components) that make the text difficult to read. There should be no unnecessary repetition in the answer. Unnecessary repetition might take the form of whole sentences that are repeated or repeated facts. The answer should be well-structured and well-organized. The answer should not just be a heap of related information, but should build from sentence to sentence to a coherent body of information about a topic.\\
\\
Evaluation Steps:\\
1. Read the entire answer thoroughly to get a general understanding of the content and structure.\\
2. Look for any dateline references or system-internal formatting errors. These issues should not be present in a coherent answer.\\
3. Examine the grammar and sentence construction for flaws such as sentence fragments, run-on sentences, or missing components that could hinder readability.\\
4. Identify any unnecessary repetition within the answer. This includes repeated sentences or redundant information that does not contribute to the progression of the text.\\
5. Assess the structure and organization of the answer. It should have a logical flow, with each sentence building upon the previous information and contributing to a comprehensive understanding of the topic.\\
6. Assign a score for coherence on a scale of 1 to 5, where 1 is the lowest and 5 is the highest based on the Evaluation Criteria.\\
\\
Answer: \textbf{\{answer\}}\\
Evaluation Form (scores ONLY):\\
- Coherence:\\
\hline
\end{longtable}

% Chinese Coherence
\begin{longtable}{|p{\textwidth}|}
    \caption{Prompt for Chinese Coherence Evaluation}
    \label{coherence_prompts_zh}\\
    \hline
    你将获得针对某个问题编写的一个答案。\\
你的任务是根据一项指标对答案进行评分。\\
请你确保仔细阅读并理解这些说明。请在审阅时保持本文档处于打开状态，并根据需要进行参考。\\
\\
评价标准：\\
连贯性 (1-5) - 所有句子的集体质量。 答案不应包含日期线、系统内部格式、大写错误或明显不合语法的句子（例如片段、缺少组件），以免文本难以阅读。 答案中不应有不必要的重复。 不必要的重复可能表现为重复整个句子或重复事实。 答案应该结构良好、组织良好。 答案不应该只是一堆相关信息，而应该逐句构建有关某个主题的连贯信息体。\\
\\
评估步骤：\\
1. 通读整个答案，大致了解内容和结构。\\
2. 查找任何日期线引用或系统内部格式错误。这些问题不应该出现在一个连贯的答案中。\\
3. 检查语法和句子结构是否存在缺陷，例如句子片段、连贯句子或可能妨碍可读性的缺失成分。\\
4. 找出答案中任何不必要的重复。这包括重复的句子或对文本的进展没有贡献的冗余信息。\\
5. 评估答案的结构和组织。它应该有一个逻辑流程，每个句子都建立在前面的信息的基础上，并有助于全面理解该主题。\\
6. 根据评估标准，按照 1 到 5 的范围为一致性打分，其中 1 为最低，5 为最高。\\
\\
回答：\textbf{\{answer\}}\\
评估形式（仅分数）：\\
- 连贯性：\\
    \hline
\end{longtable}

% English Helpfulness
\begin{longtable}{|p{\textwidth}|}
    \caption{Prompt for English Helpfulness Evaluation}
    \label{helpfulness_prompts_en}\\
    \hline
    You will be given one answer written for a question.\\
Your task is to rate the answer on one metric.\\
Please make sure you read and understand these instructions carefully. Please keep this document open while reviewing, and refer to it as needed.\\
Evaluation Criteria:\\
Helpfulness (1-5) - the degree to which the answer effectively meets the needs of the person seeking information. The answer should be articulated in a way that is easy to understand. The answer should directly address the question posed, focusing on the specific information or solution sought by the inquirer. The information provided should be correct and based on factual, verifiable data or recognized expertise. The answer includes all necessary details to fully respond to the question. It leaves no critical aspects of the question unaddressed.\\
\\
Evaluation Steps:\\
1. Begin by thoroughly reading and understanding the question asked. Identify the main points and what the inquirer is seeking.\\
2. Read the answer thoroughly to understand the information provided. Evaluate if the answer is articulated in a way that is easy to understand.\\
3. Evaluate whether the answer directly addresses the question. The information provided should be relevant to what was asked.\\
4. Ensure that the information provided is correct, factual, and based on verifiable data or recognized expertise.\\
5. Determine if the answer includes all necessary details. A helpful answer should fully respond to the question without leaving critical aspects unaddressed.\\
6. Assign a score for helpfulness on a scale of 1 to 5, where 1 is the lowest and 5 is the highest based on the Evaluation Criteria.\\
\\
Question: \textbf{\{query\}}\\
Answer: \textbf{\{answer\}}\\
Evaluation Form (scores ONLY):\\
- Helpfulness:\\
    \hline
\end{longtable}

% Chinese Helpfulness
\begin{longtable}{|p{\textwidth}|}
    \caption{Prompt for Chinese Helpfulness Evaluation}
    \label{helpfulness_prompts_zh}\\
    \hline
你将获得针对某个问题编写的一个答案。\\
你的任务是根据一项指标对答案进行评分。\\
请你确保仔细阅读并理解这些说明。请在审阅时保持本文档处于打开状态，并根据需要进行参考。\\
评价标准：\\
帮助性 (1-5) - 答案有效满足寻求信息的人的需求的程度。 答案应该以易于理解的方式表达。 答案应直接解决提出的问题，重点关注询问者寻求的具体信息或解决方案。 所提供的信息应正确且基于事实、可验证的数据或公认的专业知识。 答案包括完整回答问题的所有必要细节。 它没有遗漏问题的所有关键方面。\\
\\
评估步骤：\\
1. 首先彻底阅读并理解所提出的问题。 确定要点以及询问者想要什么。\\
2. 仔细阅读答案，理解所提供的信息。 评估答案是否以易于理解的方式表达。\\
3. 评估答案是否直接解决问题。 提供的信息应与所询问的内容相关。\\
4. 确保所提供的信息正确、真实，并且基于可验证的数据或公认的专业知识。\\
5. 确定答案是否包含所有必要的详细信息。 有用的答案应该充分回答问题，同时不遗漏关键方面。\\
6. 根据评估标准，按 1 到 5 的等级为帮助性打分，其中 1 为最低，5 为最高。\\
\\
问题：\textbf{\{query\}}\\
回答：\textbf{\{answer\}}\\
评估形式（仅分数）：\\
- 帮助性：\\
    \hline
\end{longtable}

\clearpage
% English Factuality
\begin{longtable}{|p{\textwidth}|} 
    \caption{Prompt for English Factuality Evaluation} \label{factuality_prompts_en}\\
    \hline
I will show you a question, a series of text segments, and a few reference documents. All the segments can be linked together to form a complete answer to the question. Your task is to assess whether each text segment contains factual errors with the help of the reference documents.\\
\\
Please evaluate according to the following requirements:\\
1. If the text segment only contains general introductions such as ``the method is as follows" or ``according to the material," without conveying specific information, directly judge it as ``correct."\\
2. If the text segments can find corresponding sentences in the reference document or the question as support, or can be inferred from the corresponding sentences, directly judge them as ``correct." Pay attention to the semantic consistency of key words and details.\\
3. If there are any pieces of information in the text segment that cannot be found in the reference document or the question, nor can they be inferred from the corresponding sentences, directly judge them as ``incorrect."\\
\\
Please generate the output in the following format:\\
Assessment Details: For each segment, use the reference documents and question to explain the rationale and judgment result.\\
Final Answer: List the numbers of the segments with errors (separated by commas). Please only output the numbers, do not output more details. If all information in the text segments is supported and semantically consistent with the reference documents, even if some information in the reference documents is not covered, please output ``completely correct."\\
\\
Here's an example:\\
Question: How many nuclear power plants are there in the world?\\
Segments:\\
<1>There are a total of 440 operating nuclear reactors in the world, with a total installed capacity of over 390 gigawatts (GW).\\
<2>These reactors are distributed in 30 countries around the world, with the United States owning the most reactors, followed by France, China, Japan, and Russia.\\
<3>Among them, China is building 16 new reactors with the fastest growth, followed by India with 8 under construction.\\
Reference Documents:\\
\`{}\`{}\`{}\\
\lbrack 1\rbrack Nuclear power plants operate in 32 countries, generating about one-tenth of the world's electricity. Most are located in Europe, North America, East Asia, and South Asia. The United States is the largest producer of nuclear power, while France has the largest share of electricity from nuclear power, at about 70\%. China has the fastest-growing nuclear power program, with 16 new reactors under construction, followed by India with 8 under construction.\\
\lbrack 2\rbrack As of May 2023, there are 410 operable nuclear reactors in the world, with a total electric power capacity of 368.6 gigawatts.\\
\`{}\`{}\`{}\\
\\
Here's what your output should look like:\\
Assessment Details:\\
<1>Segment 1 states: ``There are a total of 440 operating nuclear reactors in the world, with a total installed capacity of over 390 gigawatts (GW)." This information is inconsistent with the information in the reference document \lbrack 2\rbrack: ``there are 410 operable nuclear reactors in the world, with a total electric power capacity of 368.6 gigawatts." Therefore, segment 1 is incorrect.\\
<2>The information in segment 2 ``These reactors are distributed in 30 countries around the world" is inconsistent with the information in reference document \lbrack 1\rbrack: ``Nuclear power plants operate in 32 countries," and the reference document does not mention the number of nuclear reactors in ``France, China, Japan, and Russia," so segment 2 is incorrect.\\
<3>Segment 3 states: ``China is building 16 new reactors with the fastest growth, followed by India with 8 under construction." This is consistent with reference document \lbrack 1\rbrack: ``China has the fastest-growing nuclear power program, with 16 new reactors under construction, followed by India with 8 under construction." Therefore, segment 3 is correct.\\
Final Answer: 1,2\\
This means that only segments 1 and 2 contain errors, while segment 3 can find corresponding sentences in the reference document to support it and is judged to be correct.\\
\\
Below is the input:\\
Question: \textbf{\{question\}}\\
Segments: \textbf{\{segments\}}\\
Reference Documents:\`{}\`{}\`{} \textbf{\{docs\}} \`{}\`{}\`{}\\
\hline
\end{longtable}

% Chinese Factuality
\clearpage
\begin{longtable}{|p{\textwidth}|} 
    \caption{Prompt for Chinese Factuality Evaluation} \label{factuality_prompts_zh}\\
    \hline
我将向您展示一个问题、一系列文本片段和一份参考文档。所有的片段可以连起来形成对问题的完整回答。您的任务是在参考文档的帮助下评估每个文本片段是否包含事实错误。\\
\\
请按照以下要求进行评估：\\
1. 如果文本片段仅包含类似“方法如下”、“根据资料可得”这样的通用开场白而没有传递具体信息，直接判定为“正确”。\\
2. 如果文本片段都能在参考文档或者问题中找到相应句子作为支持，或者可以从相应句子推理得到，直接判定为“正确”。关注关键词和细节的语义一致性。\\
3. 如果文本片段中有任何信息在参考文档或者问题中找不到明确的支撑，也不能根据相应句子推理得到，直接判定为“错误”。\\
\\
请按照以下格式生成输出：\\
评估明细：对于每一条片段，结合参考文档或者问题说明判定理由和判定结果。\\
最终答案：列出带有错误的片段的编号（用逗号分隔）。请只输出编号，不要输出更多细节。如果所有文本片段中的信息都得到了支持且与参考文档语义一致，即使参考文档中有其他信息未被涵盖，请输出“完全正确”。\\
\\
这里有一个例子：\\
问题：全世界共有多少座核电站？\\
片段：\\
<1>世界上共有440座运行中的核电反应堆，总装机容量超过390吉瓦特（GW）。\\
<2>这些反应堆分布在世界上30个国家，美国拥有最多的反应堆，其次是法国、中国、日本和俄罗斯。\\
<3>其中，中国正在新建16座反应堆，增长最快，其次是印度，有8座在建。\\
参考文档：\\
\`{}\`{}\`{}\\
\lbrack 1\rbrack 核电站在32个国家运营，产生了世界上约十分之一的电力。大多数位于欧洲、北美、东亚和南亚。美国是最大的核能生产国，而法国的核电所占的电力比重最大，约为70\%。中国拥有增长最快的核电计划，正在建设16座新的反应堆，其次是印度，有8座在建。\\
\lbrack 2\rbrack 截至2023年5月，世界上有410座可运行的核电反应堆，总电力容量为368.6吉瓦特。\\
\`{}\`{}\`{}\\
\\
以下是您的输出：\\
评估明细：\\
<1>片段1提到：“世界上共有440座运行中的核电反应堆，总装机容量超过390吉瓦特（GW）。” 这一信息与参考文档\lbrack 2\rbrack 中的信息不一致：“世界上有410座可运行的核电反应堆，总电力容量为368.6吉瓦特。” 因此片段1是错误的。\\
<2>片段2中的信息“反应堆分布在世界上30个国家”与参考文档\lbrack 1\rbrack 中的信息不一致：“核电站在32个国家运营”，而且参考文档中没有提及“法国、中国、日本和俄罗斯”的核反应堆数量，因此片段2是错误的。\\
<3>片段3提到：“中国正在新建16座反应堆，增长最快，其次是印度，有8座在建。”与参考文档\lbrack 1\rbrack 一致：“中国拥有增长最快的核电计划，正在建设16座新的反应堆，其次是印度，有8座在建。”，因此片段3是正确的。\\
最终答案：1,2\\
这意味着只有片段1和2包含错误，而片段3都能在参考文档中找到相应句子作为支持，被判定为正确。\\
\\
以下是我的输入：\\
问题：\textbf{\{question\}}\\
片段：\textbf{\{segments\}}\\
参考文档：\`{}\`{}\`{} \textbf{\{docs\}} \`{}\`{}\`{}\\

\hline
\end{longtable}

\clearpage
\section{Prompts for Decomposition}
\label{app:prompt_decomposition}

The following are the prompts we used to decompose sentences into subclaims.

% English Decomposition
\begin{longtable}{|p{\textwidth}|}
    \caption{Prompt for English Sentence Decomposition}
    \label{segmentation_prompts_en}\\
    \hline
    Please breakdown the following sentence into independent facts: He is also a successful producer and engineer, having worked with a wide variety of artists, including Willie Nelson, Tim McGraw, and Taylor Swift.\\
- He is successful.\\
- He is a producer.\\
- He is a engineer.\\
- He has worked with a wide variety of artists.\\
- Willie Nelson is an artist.\\
- He has worked with Willie Nelson.\\
- Tim McGraw is an artist.\\
- He has worked with Tim McGraw.\\
- Taylor Swift is an artist.\\
- He has worked with Taylor Swift.\\
\\
Please breakdown the following sentence into independent facts: Michael Collins (born October 31, 1930) is a retired American astronaut and test pilot who was the Command Module Pilot for the Apollo 11 mission in 1969.\\
- Michael Collins was born on October 31, 1930.\\
- Michael Collins is retired.\\
- Michael Collins is an American.\\
- Michael Collins was an astronaut.\\
- Michael Collins was a test pilot.\\
- Michael Collins was the Command Module Pilot.\\
- Michael Collins was the Command Module Pilot for the Apollo 11 mission.\\
- Michael Collins was the Command Module Pilot for the Apollo 11 mission in 1969.\\
\\
Please breakdown the following sentence into independent facts: He was an American composer, conductor, and musical director.\\
- He was an American.\\
- He was a composer.\\
- He was a conductor.\\
- He was a musical director.\\
\\
Please breakdown the following sentence into independent facts: In 1970, the Empire State Building in New York City was the tallest building in the United States and the world, standing at 1,250 feet tall.\\
- The Empire State Building is in New York City.\\
- In 1970, the Empire State Building was the tallest building in the United States.\\
- In 1970, the Empire State Building was the tallest building in the world.\\
- The Empire State Building stands at 1,250 feet tall.\\
\\
Please breakdown the following sentence into independent facts: The Willis Tower (formerly the Sears Tower) in Chicago was the first to do so, reaching 1,450 feet in 1973.\\
- The Willis Tower is formerly called the Sears Tower.\\
- The Willis Tower is in Chicago.\\
- The Willis Tower reached 1,450 feet in 1973.\\
\\
Please breakdown the following sentence into independent facts: The current tallest building in the United States is One World Trade Center in New York City, which stands at 1,776 feet.\\
- The current tallest building in the United States is One World Trade Center.\\
- One World Trade Center is in New York City.\\
- One World Trade Center stands at 1,776 feet.\\
\\
Please breakdown the following sentence into independent facts: William E. Moerner is an American physical chemist who was affiliated with the University of Sussex as a visiting professor.\\
- William E. Moerner is an American.\\
- William E. Moerner is an physical chemist.\\
- William E. Moerner was affiliated with the University of Sussex.\\
- William E. Moerner was affiliated with the University of Sussex as a visiting professor.\\
\\
Please breakdown the following sentence into independent facts: Sir Harold Walter Kroto, an English chemist, shared the 1996 Nobel Prize in Chemistry with Robert Curl and Richard Smalley for their discovery of a new form of carbon, buckminsterfullerene, also known as buckyballs.\\
- Sir Harold Walter Kroto is English.\\
- Sir Harold Walter Kroto is an chemist.\\
- Sir Harold Walter Kroto won the Nobel Prize in 1996.\\
- Sir Harold Walter Kroto won the Nobel Prize in Chemistry.\\
- Sir Harold Walter Kroto shared the Nobel Prize with Robert Curl and Richard Smalley.\\
- They won the prize for their discovery of a new form of carbon, buckminsterfullerene, also known as buckyballs.\\
\\
Please breakdown the following sentence into independent facts: \textbf{\{sentence\}}\\
    \hline
\end{longtable}

% Chinese Decomposition
\begin{longtable}{|p{\textwidth}|}
    \caption{Prompt for Chinese Sentence Decomposition}
    \label{segmentation_prompts_zh}\\
    \hline
    请将下面的句子分解为独立的陈述：王羲之是东晋时期的书法家，被誉为书圣，其代表作《兰亭序》被认为是楷书的巅峰之作。\\
- 王羲之是东晋时期的人物。\\
- 王羲之是书法家。\\
- 王羲之被誉为书圣。\\
- 王羲之的代表作是《兰亭序》。\\
- 《兰亭序》被认为是楷书的巅峰之作。\\
\\
请将下面的句子分解为独立的陈述：北京住房公积金缴存人购买北京市行政区域内商品住房、经济适用房及限价商品房、公有住房及集资合作建房、危改拆迁征收回迁房，可申请提取住房公积金。\\
- 北京住房公积金缴存人购买北京市行政区域内商品住房可申请提取住房公积金。\\
- 北京住房公积金缴存人购买北京市行政区域内经济适用房可申请提取住房公积金。\\
- 北京住房公积金缴存人购买北京市行政区域内限价商品房可申请提取住房公积金。\\
- 北京住房公积金缴存人购买北京市行政区域内公有住房可申请提取住房公积金。\\
- 北京住房公积金缴存人购买北京市行政区域内集资合作建房可申请提取住房公积金。\\
- 北京住房公积金缴存人购买北京市行政区域内危改拆迁征收回迁房可申请提取住房公积金。\\
\\
请将下面的句子分解为独立的陈述：中国的茶文化有着几千年的历史，涵盖了茶叶的种植、加工、饮用习俗以及茶艺表演等方面，是中国传统文化中不可分割的一部分，也对周边国家如日本和韩国的茶文化产生了深远的影响。\\
- 中国的茶文化有几千年的历史。\\
- 茶文化包括茶叶种植。\\
- 茶文化包括茶叶加工。\\
- 茶文化包括茶叶饮用习俗。\\
- 茶文化包括茶艺表演。\\
- 茶文化是中国传统文化中不可分割的一部分。\\
- 日本是中国的周边国家\\
- 韩国是中国的周边国家\\
- 中国的茶文化对日本的茶文化有深远影响。\\
- 中国的茶文化对韩国的茶文化有深远影响。\\
\\
请将下面的句子分解为独立的陈述：布洛芬属于非甾体类解热镇痛药，它通过抑制前列腺素的合成从而发挥解热、镇痛的作用，从而达到使发热者退热、缓解疼痛带来的不适感等作用。\\
- 布洛芬是非甾体类解热镇痛药。\\
- 布洛芬通过抑制前列腺素的合成来发挥作用。\\
- 布洛芬具有解热的作用。\\
- 布洛芬具有镇痛的作用。\\
- 布洛芬能帮助发热者退热。\\
- 布洛芬能缓解疼痛带来的不适感。\\
\\
请将下面的句子分解为独立的陈述：屠呦呦是一位杰出的中国药理学家，以其发现抗疟疾药物青蒿素而闻名于世，这一发现源于对中国古代药书的深入研究，对全球抗击疟疾的努力作出了历史性的贡献，她的工作不仅挽救了无数生命，还因此荣获2015年诺贝尔生理学或医学奖。\\
- 屠呦呦是杰出的。\\
- 屠呦呦是中国人。\\
- 屠呦呦是药理学家。\\
- 屠呦呦因发现青蒿素而成为世界闻名。\\
- 青蒿素是抗疟疾药物。\\
- 青蒿素的发现是基于对中国古代药书的研究。\\
- 屠呦呦的研究对全球抗击疟疾做出了重大贡献。\\
- 青蒿素的发现挽救了无数人的生命。\\
- 屠呦呦获得了2015年诺贝尔生理学或医学奖。\\
\\
请将下面的句子分解为独立的陈述：中国的故宫是明清两代的皇宫，占地超过72万平方米，包含了约8000间房屋，拥有丰富的历史文物，每年吸引数百万游客前来参观，是世界上现存规模最大、保存最为完整的木质结构古建筑群之一。\\
- 故宫是中国的。\\
- 故宫曾是明代的皇宫。\\
- 故宫曾是清代的皇宫。\\
- 故宫占地超过72万平方米。\\
- 故宫包含约8000间房屋。\\
- 故宫内拥有丰富的历史文物。\\
- 每年有数百万游客参观故宫。\\
- 故宫是世界上最大的木质结构古建筑群之一。\\
- 故宫的保存状况非常完整。\\
\\
请将下面的句子分解为独立的陈述：袁隆平，被称为“杂交水稻之父”，是中国著名的农学家，其领导的科研团队成功开发出了第一代杂交水稻，大幅提高了水稻产量，他的研究成果被广泛应用于亚洲、非洲等多个国家和地区，极大地缓解了这些地区的饥饿问题，袁隆平本人因此获得了包括“世界粮食奖”在内的众多国际荣誉，被世界各国誉为人类粮食生产的英雄。\\
- 袁隆平被称为“杂交水稻之父”。\\
- 袁隆平是中国人。\\
- 袁隆平是著名的农学家。\\
- 袁隆平领导的科研团队开发出了第一代杂交水稻。\\
- 杂交水稻的开发显著提升了水稻产量。\\
- 杂交水稻技术被应用于亚洲多个国家和地区。\\
- 杂交水稻技术被应用于非洲多个国家和地区。\\
- 袁隆平的工作帮助缓解了世界饥饿问题。\\
- 袁隆平获得了包括“世界粮食奖”在内的多项国际荣誉。\\
- 袁隆平被世界各国誉为人类粮食生产的英雄。\\
\\
请将下面的句子分解为独立的陈述：敦煌是中国甘肃省的一个历史名城，丝绸之路的重要节点，以其保存下来的一系列壁画和雕塑闻名于世，其中最著名的莫过于敦煌莫高窟，这些艺术作品不仅展示了古代丝绸之路的繁荣，也反映了不同宗教文化的交融，敦煌成为了研究中国古代历史、宗教和文化交流的宝贵资料。\\
- 敦煌在中国甘肃省。\\
- 敦煌是历史名城。\\
- 敦煌是丝绸之路的重要节点。\\
- 敦煌以壁画闻名于世。\\
- 敦煌以雕塑闻名于世。\\
- 敦煌莫高窟是敦煌最著名的。\\
- 莫高窟的艺术作品展示了古代丝绸之路的繁荣。\\
- 敦煌艺术反映了不同宗教文化的交融。\\
- 敦煌是研究中国古代历史的宝贵资料。\\
- 敦煌是研究中国古代宗教的宝贵资料。\\
- 敦煌是研究中国古代文化交流的宝贵资料。\\
\\
请将下面的句子分解为独立的陈述：\textbf{\{sentence\}}\\
    \hline
\end{longtable}

\section{Dataset Modification Samples} \label{dataset_answer_improvement}

We samples query, reference, and answers from WebGLM-QA (English-only) and WebCPM (Chinese-only) and compare them with the answer constructed using our outline-enhanced technique. Answers in our dataset has organization pattern, the outline and the answer. It is clear that our answers are longer and have a better logic structure. 

\begin{longtable}[h]{p{0.05\textwidth}p{0.25\textwidth}p{0.25\textwidth}p{0.35\textwidth}}
\caption{Examples for Dataset Modification}
\label{tab:dataset_modification} \\
\hline
\textbf{Query} & \textbf{References} & \textbf{Answer in Original Dataset} & \textbf{Answer in Our Dataset} \\
\hline

why aren't all tin cans ring pull??&
[1]Thanks to @alanwood01, we think we’ve been using ring pulls on tin cans in completely the wrong way. But what do you think?

[2]Thanks to @alanwood01, we think we’ve been using ring pulls on tin cans in completely the wrong way. But what do you think?

[3]They're not 100\% ideal in all circumstances. Soup cans with ring pulls must have much thinner lids to work, which means the lids are more vulnerable to damage. As an educated guess, I would imagine that the ring pull lids cost slightly more to manufacture, too, which isn't at all a small consideration when you're talking about something as low-margin as most groceries.

[4]Ring pulls are expensive compared to the standard lid. It's fractions of a penny, but it adds up when you manufacture a few million units in cans a day.

[5]``In my civilized European country all cans are ring-pull, even the ones with tuna. OP must be a primitive American, where you have to use stupid can openers for everything.&
Ring pull cans are not used universally because they are more expensive to manufacture compared to the standard lid[3], and the soup can lids must be thinner to work, making them more vulnerable to damage[3]. In addition, the cost difference adds up when manufacturing a few million cans a day[4], making the ring pull lids less cost-effective.[3]
&[Structure]:

Cause-Effect

[Outline]:

1. Manufacturing cost considerations (based on [3] and [4])

2. Practicality in terms of product safety and integrity (based on [3])

[Answer]:

The absence of ring pulls on all tin cans can be attributed primarily to two factors: cost concerns and practicality related to product safety.

\#\#\# Manufacturing Cost Considerations

Ring pulls, while convenient for consumers, are more expensive to produce than standard lids. This cost difference, although seemingly minor on a per-unit basis, becomes significant when considering the scale of production. Manufacturers often produce millions of cans daily, and even fractions of a penny per can add up to a considerable amount. As groceries are generally low-margin products, companies are inclined to minimize production costs wherever possible to maintain profitability.

\#\#\# Practicality for Product Safety

In addition to cost implications, the feasibility of using ring pulls on certain products is limited by the need for thinner lids. This requirement for a thinner lid makes the cans more vulnerable to damage, posing a risk to the integrity of the product inside. Soup cans, for example, must maintain their structure to ensure safe storage and transportation. The decision to forgo ring pulls on some tin cans is a measure to prevent potential damage and the associated waste or customer dissatisfaction that could result from compromised product packaging.\\ \hline 
Why don't surfers wear life vests and helmets?&
[1]One of the major reasons surfers don't use life vests is that when they are in the water they are strapped to a giant floatation device, their surfboard.

[2]Other than the fact that they look bad, the main reason why surfers don’t wear life jackets is that they get in the way. With a PFD on, it will be harder to paddle and duck dive. Furthermore, surfers are literally standing on a big PFD – their surfboards – which are attached to their legs with a rope.

[3]Let's be honest here, this is probably the \#1 reasons apart from the “we don't really need it” reason that surfers don't wear life jackets.

[4]Now, arguably the main reason why most surfers don’t wear life jackets is that they get in the way both in terms of fun and, ironically, in the way of safety as well.

[5]While surfers don’t wear life vests (there are specific requirements for a device to be considered life saving), it’s not a surprise that big wave surfers wear special inflatable PFDs.&
Surfers generally don't wear life vests and helmets because they are strapped to a giant floatation device, their surfboard[1], and life jackets get in the way of paddling and duck diving[2][4]. Furthermore, surfers don't need to wear life jackets because they are standing on a big PFD - their surfboards - which are attached to their legs with a rope[2]. For bigger waves, some surfers wear special inflatable PFDs[5].
&[Structure]:

Comparative

[Outline]:

1. Surfboard as a flotation device (based on [1])

2. Hindrance to movement and safety (based on [2] and [4])

3. Aesthetic and traditional reasons (based on [3])

4. Exception for big wave surfers (based on [5])

[Answer]:

Surfers typically eschew traditional life vests and helmets due to a combination of practical, safety, and cultural reasons, as well as the nature of their equipment.

\#\#\# Surfboard as a Flotation Device

A primary reason surfers don't utilize life vests is their surfboard, which essentially acts as a large flotation device. Since surfboards are buoyant and tethered to the surfer by a leash, they provide significant flotation support without the need for an additional life vest.

\#\#\# Hindrance to Movement and Safety

Life vests are generally avoided because they can interfere with the surfer's ability to move freely. The requirement to paddle effectively and perform duck dives — a maneuver used to dive beneath oncoming waves — is hampered by the bulk of a personal flotation device (PFD). Ironically, rather than increasing safety, a life vest can actually present a hazard by restricting mobility in the water.

\#\#\# Aesthetic and Traditional Reasons

The avoidance of life vests is also influenced by aesthetics and tradition within the surfing community. Surfers often prefer the freedom of not wearing what may be considered unattractive gear, adhering to a longstanding cultural norm of surfing without excessive equipment.

\#\#\# Exception for Big Wave Surfers

While standard practice is to not wear life vests, there is an exception for big wave surfers, who may use specialized inflatable PFDs. These devices are designed with the specific risks of big wave surfing in mind, balancing the need for flotation with the necessity of not hindering movement.

In conclusion, the decision not to wear life vests and helmets while surfing is largely based on the surfboard's role as a flotation aid, the importance of unhindered movement for both performance and safety, cultural factors regarding appearance, and exceptions made for the unique conditions faced by big wave surfers.\\ \hline
.NET是什么？ &
[1].NET是一种用于构建多种应用的免费开源开发平台，可以使用多种语言，编辑器和库开发Web应用、Web API和微服务、云中的无服务器函数、云原生应用、移动应用、桌面应用、Windows WPF、Windows窗体、通用 Windows平台 (UWP)、游戏、物联网 (IoT)、机器学习、控制台应用、Windows服务。.NET类库在不同应用和应用类型中共享功能，无论构建哪种类型的应用，代码和项目文件看起来都一样，可以访问每个应用的相同运行时、API和语言功能。&
.NET是一种用于构建多种应用的免费开源开发平台，可以使用多种语言，编辑器和库开发Web应用、Web API和微服务、云中的无服务器函数、云原生应用、移动应用、桌面应用、Windows WPF、Windows窗体、通用 Windows平台 (UWP)、游戏、物联网 (IoT)、机器学习、控制台应用、Windows服务。.NET类库在不同应用和应用类型中共享功能，无论构建哪种类型的应用，代码和项目文件看起来都一样，可以访问每个应用的相同运行时、API和语言功能。&【结构】：

并列

【提纲】：

1. .NET的定义与性质（仅使用[1]回答）

2. .NET的开源特性（仅使用[1]回答）

3. .NET的应用领域（仅使用[1]回答）

4. .NET的编程语言支持与API（仅使用[1]回答）

【回答】：

.NET是一种多功能的开发平台，旨在支持开发人员构建各种类型的应用程序。它具有多种特点和功能，下面进行详细介绍：

1. .NET的定义与性质

.NET是一个免费且开源的开发平台，它提供了一套工具和资源，使开发者能够使用多种编程语言、编辑器和库来构建应用程序。

2. .NET的开源特性

作为一个开源平台，.NET允许开发者自由地访问其源代码，共同贡献和改进平台的功能和性能。

3. .NET的应用领域

该平台适用于多种应用程序的开发，包括但不限于Web应用程序、Web API和微服务、无服务器函数、云原生应用、移动应用、桌面应用、游戏、物联网 (IoT) 设备、机器学习项目以及控制台应用程序等。

4. .NET的编程语言支持与API

.NET支持多种编程语言，这意味着开发者可以根据自己的偏好选择合适的语言来构建项目。此外，.NET提供了丰富的API和类库来促进不同应用程序之间功能的共享，无论是哪种类型的应用，其代码和项目文件在结构上都保持一致性，开发者可以访问相同的运行时环境、API和语言功能，这极大地便利了开发过程和维护工作。
\\ \hline 
狗如何理解人类语言？&
[1]狗狗对于铲屎官的语言本身会进行思考和理解，但也会观察铲屎官说话时的表情和动作来帮助解读，这就说明咱们低估了狗狗的学习能力。
而对于狗狗能听懂人话还有另一个解释，那就是条件反射。不错，就拿上面提到的给狗狗起名字以及一些常用的指令，通过反复记忆狗狗就能形成条件反射记住这些词汇的含义。
每种生物都有属于自己的那套语言功能系统，人与狗之间的语言系统明显是不同的，那么想要做到完全了解彼此的

[2]话语肯定是不太可能的。
 不过虽然狗狗不能完全明白铲屎官的对话，却可以通过情绪来辨别话语的意思，说到底这也是狗狗经过数千年的驯化后与人类达成的默契，它们并不只是依靠条件反射来读懂铲屎官，更多的时候它们也在思考，也在努力走进铲屎官的世界。
 
[3]虽然不同犬类对理解人类语言的水平和动机会有不同，但当它们听到学过的单词时，脑部都会产生神经反应而不只是巴甫洛夫的条件反射。本次研究的结果不能说明语言是和狗沟通最有效的方式，真正有效的其实是犬类的神经奖励机制，才能让训狗更加有效。而人类更倾向于通过语言来训练狗，这是从人类的认知角度出发得到的主观性常识，实际上对于犬类来说视觉感官才更加有效。&
狗对人类语言的理解主要从两个角度来说，一是狗听到学过的单词时，脑部都会产生神经反应而不只是巴甫洛夫的条件反射。狗狗对于人类的语言本身会进行思考和理解，但也会观察人类说话时的表情和动作来帮助解读。二是条件反射，通过反复记忆狗狗就能形成条件反射记住这些词汇的含义。&【结构】：

因果

【提纲】：

1. 狗狗的语言理解包括语言和非语言因素（仅使用[1]回答）

2. 狗狗对于人类话语的理解与情绪辨识相关（仅使用[2]回答）

3. 狗狗理解人类语言的神经机制（仅使用[3]回答）

【回答】：

狗狗理解人类语言的过程是一个复杂的认知活动，涉及到多种因素和机制。

首先，狗狗不仅仅是通过听觉接受人类的语言，它们还会结合人类的表情和动作来辅助理解。这表明狗狗具备一定的学习能力，能够在一定程度上思考和理解它们的主人所表达的内容。同时，狗狗也会通过条件反射机制来记住和响应主人的命令和名字。

其次，狗狗能够通过感知人类的情绪来辨别话语的意义，这是数千年驯化过程中形成的一种与人类的沟通默契。在和人类的互动过程中，它们并不完全依赖条件反射，而是通过观察和思考来更深入地理解人类的世界。

最后，科学研究表明，狗狗在听到熟悉的单词时会在大脑中产生神经反应，这不仅仅是简单的条件反射。有效的犬类训练实际上和它们的神经奖励机制有关，而非单纯的语言训练。人类可能更倾向于用语言来训练狗，但从狗的认知角度看，视觉感官可能是更有效的沟通方式。

综上所述，狗狗理解人类语言的能力是多方面的，涉及到语言的听觉处理、非语言的情绪和肢体语言认知，以及大脑的神经奖励机制。这些因素共同作用，让狗狗能够更好地理解和响应人类的指令和情感。
\\ \hline
\end{longtable}

\clearpage

\section{Generation Improvement Samples} \label{example_our_model_baseline}

We provide examples from the test set of WebGLM-QA (in English) and WebCPM (in Chinese). For each test set, we compare the open sourced method with best performance on that dataset with our FoRAG-L 7B. 

\begin{longtable}[h]{p{0.05\textwidth}p{0.25\textwidth}p{0.25\textwidth}p{0.35\textwidth}}
\caption{Examples for Generation Improvement}
\label{tab:WebGLM}\\
\hline \\
\textbf{Query} & \textbf{References} & \textbf{Answer from SOTA Model} & \textbf{Answer from Our Best Model}\\
\hline
IPod和磁带播放机的工作原理？ & [1]ipod video的芯片是WM8758，touch的是WM8758BG，似乎是多了两个字母后缀。中间两个是闪存片，右边的是微处理器。
在MP3播放器中，解码芯片将存储在闪存或者硬盘里面的数据格式歌曲解码成数字信号，再通过模拟输出电路转化为模拟信号，并通过耳机播放出来”
这个数模转换的过程，和“内核”是无关的，“内核”只输出数字信号而已，很显然“解码芯片”并不是三星、西格玛、portal player之类的“内核” 。

[2]在音频领域，磁带（以小型磁带的形式）是分发音乐的流行方式。
磁带本身实际上非常简单。它由薄的塑料基材组成，并与该基底粘合，是一种氧化铁粉末涂层。氧化物通常与粘合剂混合以将其附着到塑料上，并且还包括某种干润滑剂以避免磨损记录器。
氧化铁（FeO）是我们常见的红锈。氧化铁（Fe 2 O 3）是铁的另一种氧化物。磁赤铁矿或氧化铁是该物质的通用名称。
这种氧化物是一种铁磁性材料，这意味着如果将其暴露在磁场中，它会被磁场永久磁化。这种能力赋予磁带两个最吸引人的功能：
您可以立即录制任何想要的内容，录像带会记住您录制的内容，以便随时播放。
您可以随时擦除磁带并在其上记录其他内容。
 & IPod和磁带播放机的工作原理是通过解码芯片将存储在闪存或者硬盘里面的数据格式歌曲解码成数字信号，再通过模拟输出电路转化为模拟信号，并通过耳机播放出来。这种数模转换的过程，和“内核”是无关的，“内核”只输出数字信号而已，很显然“解码芯片”并不是三星、西格玛、portal player之类的“内核” 。此外，磁带本身实际上非常简单，它由薄的塑料基材组成，并与该基底粘合，是一种氧化铁粉末涂层。氧化物通常与粘合剂混合以将其附着到塑料上，并且还包括某种干润滑剂以避免磨损记录器。这种氧化物是一种铁磁性材料，这意味着如果将其暴露在磁场中，它会被磁场永久磁化。这种能力赋予磁带两个最吸引人的功能：您可以立即录制任何想要的内容，录像带会记住您录制的内容，以便随时播放；您可以随时擦除磁带并在其上记录其他内容。这两个功能使磁带和磁盘如此受欢迎，它们是即时的，并且很容易更改。 &【结构】：

并列

【提纲】：

1. IPod和磁带播放机的芯片功能（仅使用[1]回答）

2. 磁带播放机的物理结构（仅使用[2]回答）

【回答】：

IPod和磁带播放机的工作原理涉及到芯片的解码功能以及磁带的物理结构。

\#\#\# IPod和磁带播放机的芯片功能

在IPod播放机上，芯片的功能是解码音频数据。IPod Video使用的是WM8758芯片，而Touch版本则使用WM8758BG版本，这两者之间的区别在于闪存和处理器的组合。解码芯片负责将音乐文件，以及其他数据格式的歌曲解码成数字信号。这一过程涉及到模拟输出电路的转换，并通过耳机播放出来。这个数字到模拟转换过程并不涉及芯片本身的内核功能，而是额外的解码过程。

\#\#\# 磁带播放机的物理结构

磁带播放机的物理结构相对简单。磁带是由塑料基材组成，覆盖有一层氧化铁粉末涂层，这是一种铁磁性材料。在磁场中，氧化铁会被永久磁化，这是磁带记录音乐和视频的原理。磁带可以被立即录制和擦除内容，从而允许用户快速更改媒体内容。这种即时性和容易更改的特点是磁带播放机受欢迎的原因之一。 
\\ \hline 
什么限制了地球上生物的大小？ & [1]至少以地球的标准看，限制动物体型的因素主要有三个：
1、气温；
2、大气含氧量。
3、动物自身生理构造因素：主要是呼吸系统、循环系统的效能会对体型有很大限制。
气温其实主要对植物有重大影响，气温高，植物生长快，动物会有更多食物，容易长得更大。
根据地质历史温度与巨型动物的关系的研究表明，总的来说气温高有利于动物体型增大（虽然没有大气含氧量的影响大）。
大气含氧量和动物自身呼吸/循环系统效能决定了动物新陈代谢的速率。
地球历史上，气温高、大气含氧量高的时代，动物体型也普遍偏大，例如古生代某些时期的巨型昆虫等等，现代地球已经养不活那样大的昆虫了。 
 & 地球上生物的大小受到气温、大气含氧量和动物自身生理构造因素的限制。首先，气温对植物有重大影响，气温高，植物生长快，动物会有更多食物，容易长得更大；其次，大气含氧量决定了动物新陈代谢的速率，而动物自身呼吸/循环系统效能也会对体型有很大限制；最后，根据地质历史温度与巨型动物的关系的研究表明，总的来说气温高有利于动物体型增大（虽然没有大气含氧量的影响大）。因此，气温、大气含氧量和动物自身生理构造因素共同限制了地球上生物的大小。 & 【结构】：
总分总

【提纲】：

1. 气温对生物体型的影响（仅使用[1]回答）

2. 大气含氧量对生物体型的影响（仅使用[1]回答）

3. 生理构造因素对生物体型的影响（仅使用[1]回答）

【回答】：

地球上生物的大小受到多种因素的限制，其中气温、大气含氧量和生物自身的生理构造因素是主要的限制因素。

\#\#\# 气温对生物体型的影响

气温是影响生物体型的重要因素，尤其对植物有着显著的影响。当气温较高时，植物能更快地生长，这意味着动物就有了更多的食物来源，从而可能长得更大。地球历史上，高温时期通常伴随着大型生物的出现和繁衍。

\#\#\# 大气含氧量对生物体型的影响

大气中的含氧量也对生物的体型有重要影响。高大气含氧量意味着生物能够更有效地进行新陈代谢，这会加速生物体型的增大。古生代某些时期的地球环境具有较高的气温和大气含氧量，这导致了当时出现了巨型生物的现象。而随着地球气候的温和化，现代地球上已经无法养活像古生代的巨型昆虫了。

\#\#\# 生理构造因素对生物体型的影响

生物的自身生理构造，如呼吸系统和循环系统的效能，也会对其体型有很大的限制。一个有效的呼吸系统和循环系统能够提供足够的氧气和营养物质，以支持生物体型的增长。

综上所述，气温、大气含氧量和生物自身的生理构造因素共同限制了地球上生物的大小。气温对植物生长速率有着直接的影响，而大气含氧量和生物的生理系统效能则影响了新陈代谢速率，这些因素共同决定了生物体型的最大范围。 \\ \hline
Why do iced coffee drinks have more caffeine than hot ones? & [1]While it’s hard to pin down the precise amount of caffeine in any given iced coffee, we can make several assumptions based on what we know of caffeine. Since caffeine is soluble in water, hot water remains the best way to extract caffeine from coffee beans, giving iced coffee an edge over cold brew due to the brewing method.

But, iced coffee brewed with hot water also tends to have a lower coffee to water ratio that somewhat dilutes the caffeine content, as does the ice you need to chill the coffee.

[2]Iced coffee can be made with leftover hot coffee, and will have more caffeine in it than a cold brew coffee. Cold-brew coffees can have a lot of caffeine in them if they are highly concentrated, but the amount of caffeine that can be concentrated down into the cold brew depends on how the coffee was brewed initially.

[3]ICED COFFEE: A typical 16-oz. iced coffee has ~165 mg of caffeine.

COLD BREW: A typical 16-oz. cold brew has ~200 mg of caffeine. So, in the simplest of terms, cold brew has more caffeine than iced coffee (and, depending on how it’s brewed sometimes it can be nearly double that of its cold coffee counterpart!) But, why is there such a huge difference in the caffeine contents of these drinks?

THE BREW SAYS IT ALL

It all comes down to how each one is brewed. Iced coffee is brewed as regular coffee, which then goes through a cooling process and is served over ice. Cold brew, on the other hand, is created by steeping ground coffee at room temperature and then filtering out the grounds. So, it’s essentially the most concentrated version of coffee you can get. Once you filter out the grounds, the concentrate is combined with water for (generally) a 50/50 coffee concentrate/water split, and TA-DA! You have cold brew.
 & There is more caffeine in iced coffee than in hot coffee due to the brewing method [1, 2, 3]. Iced coffee is brewed as regular coffee, which then goes through a cooling process and is served over ice [1]. Cold brew, on the other hand, is created by steeping ground coffee at room temperature and then filtering out the grounds [1]. Once you filter out the grounds, the concentrate is combined with water for (generally) a 50/50 coffee concentrate/water split [1].  Iced coffee brewed with hot water also tends to have a lower coffee to water ratio that somewhat dilutes the caffeine content, as does the ice you need to chill the coffee [2]. & [Structure]:

Comparative-Cause-Effect

[Outline]:

1. Brewing method and extraction efficiency (based on [1] and [3])

2. Coffee to water ratio and ice dilution (based on [1])

3. Concentration of caffeine in cold brew (based on [2] and [3])

[Answer]:

iced coffee drinks contain more caffeine than their hot counterparts primarily due to differences in the brewing process and the resulting concentration of caffeine.

\#\#\# Brewing Method and Extraction Efficiency

The extraction of caffeine from coffee beans is most efficient when using hot water, as caffeine is soluble in water. This is the reason why iced coffee, which is often made by pouring hot coffee over ice, tends to have a higher caffeine content compared to cold brew coffee. However, the caffeine content can be diluted in iced coffee due to the use of hot water in the brewing process and the inclusion of ice which requires water to chill the drink.

\#\#\# Coffee to Water Ratio and Ice Dilution

When making iced coffee, the coffee to water ratio is typically higher than in cold brew coffee. This is because iced coffee is made by brewing hot coffee and then cooling it down, which results in a lower concentration of caffeine per serving. Conversely, cold brew coffee is brewed at room temperature and then filtered to remove the grounds, concentrating the caffeine.

\#\#\# Concentration of Caffeine in Cold Brew

Cold brew coffee can have a significantly higher caffeine content than iced coffee. This is because the cold brew process involves steeping ground coffee at room temperature and filtering it to create a highly concentrated coffee concentrate. Typically, a 16-ounce cold brew contains about 200 mg of caffeine, compared to iced coffee which may have around 165 mg of caffeine. The difference in caffeine content can sometimes be nearly double, depending on the specific brewing methods used for each drink.

In summary, the higher caffeine content in iced coffee drinks as opposed to hot coffees is attributed to the brewing techniques that extract more caffeine into the beverage, the dilution effect from using ice, and the concentration of caffeine in the cold brew process. \\ \hline 
Who are the Kurdish people? & [1]Kurds speak the Kurdish languages and the Zaza–Gorani languages, which belong to the Western Iranian branch of the Iranian languages in the Indo-European language family.[33][34]

[2]Kurds (Kurdish: Kurd) or Kurdish people are an Iranian[28][29][30] ethnic group native to the mountainous region of Kurdistan in Western Asia, which spans southeastern Turkey, northwestern Iran, northern Iraq, and northern Syria.[31] There are exclaves of Kurds in Central Anatolia, Khorasan, and the Caucasus, as well as significant Kurdish diaspora communities in the cities of western Turkey (in particular Istanbul) and Western Europe (primarily in Germany). The Kurdish population is estimated to be between 30 and 45 million.[2][32]

[3]The Kurds are one of the indigenous peoples of the Mesopotamian plains and the highlands in what are now south-eastern Turkey, north-eastern Syria, northern Iraq, north-western Iran and south-western Armenia.

Today, they form a distinctive community, united through race, culture and language, even though they have no standard dialect. They also adhere to a number of different religions and creeds, although the majority are Sunni Muslims.

[4]The Kurdish people are an ethnic minority group without an official state. Before World War I, Kurds lived a nomadic lifestyle until the breakup of the Ottoman Empire, which stripped them of their freedom and divided them across several nation states.

Today, there are an estimated 25-30 million Kurds, the majority living in a region that stretches across parts of Turkey, Iraq, Iran, Syria and Armenia.

Most Kurds are Sunni Muslims, but the Kurdish population has diverse cultural, social, religious and political traditions as well as a variety of dialects.
 & The Kurds are an Iranian ethnic group native to the mountainous region of Kurdistan in Western Asia, which spans southeastern Turkey, northwestern Iran, northern Iraq, and northern Syria. [1][2] They are one of the indigenous peoples of the Mesopotamian plains and the highlands in what are now south-eastern Turkey, north-eastern Syria, northern Iraq, north-western Iran and south-western Armenia. [3]

Kurds speak the Kurdish languages and the Zaza–Gorani languages, which belong to the Western Iranian branch of the Iranian languages in the Indo-European language family. [1] They are an ethnic minority group without an official state. Before World War I, Kurds lived a nomadic lifestyle until the breakup of the Ottoman Empire, which stripped them of their freedom and divided them across several nation states. [4] Today, there are an estimated 25-30 million Kurds, the majority living in a region that stretches across parts of Turkey, Iraq, Iran, Syria and Armenia. [4] Most Kurds are Sunni Muslims, but the Kurdish population has diverse cultural, social, religious and political traditions as well as a variety of dialects. [4] & [Structure]:

General-Specific-General

[Outline]:

1. Ethnic and linguistic identity (based on [1] and [3])

2. Geographical distribution and diaspora (based on [2])

3. Cultural and religious characteristics (based on [3] and [4])

[Answer]:

The Kurdish people are an ethnic group with a rich cultural and linguistic heritage, native to the region known as Kurdistan in Western Asia.

\#\#\# Ethnic and Linguistic Identity

Kurds speak Kurdish languages and Zaza–Gorani languages, which are part of the Western Iranian branch of the Iranian languages within the Indo-European family. This linguistic connection underscores their unique identity, both as an ethnic group and as a linguistic minority.

\#\#\# Geographical Distribution and Diaspora

Geographically, Kurds are spread across a region that includes parts of Turkey, Iraq, Iran, Syria, and Armenia. Their historical homeland, known as Kurdistan, encompasses significant exclaves in Central Anatolia, Khorasan, and the Caucasus. In addition to their native regions, Kurdish diaspora communities are notable in western Turkey, particularly Istanbul, and in Western Europe, with a substantial presence in Germany.

\#\#\# Cultural and Religious Characteristics

Culturally, Kurds share a common identity through race, language, and religion, even though they do not have a standard dialect. The majority of Kurds today adhere to Sunni Islam, reflecting a common religious practice among the community. Despite this religious affiliation, Kurdish culture is diverse, encompassing various social, religious, political traditions, and dialects. 

Before the dissolution of the Ottoman Empire following World War I, Kurds led a nomadic life. The subsequent division of the Ottoman Empire into several nation-states resulted in the loss of Kurdish freedom and led to a complex political situation where Kurds found themselves spread across several countries, without an official state of their own.

In summary, the Kurdish people are characterized by their distinct ethnic and linguistic background, their geographic distribution and diaspora, and the rich tapestry of their cultural and religious practices. Despite facing historical challenges, they maintain a strong sense of identity and community. \\ \hline
Why do we use shaving plugs in bathrooms, are they safer that normal 3 pin plugs? & [1]UK building regulations mean you cannot fit a regular 3 pin mains power socket in a bathroom, even the ‘shaver socket’ has to be a safe distance away from the bath or shower. This is often why many sockets are part of a light fitting above a mirror and hidden behind a little door, making it harder to access. All to protect you and I from injuring ourselves.

Other countries around the world may have normal sockets in their bathrooms. Why? For some it is because of the lower voltage used, so the effects of electric shock and serious injury are reduced. In other cases it is less established safety measures and in others there is a trust that you as an individual won’t do anything silly enough to put your life at risk.

[2]A bathroom shaver socket is a specialized electrical socket designed for safe use of electrical shavers in the bathroom. This socket design is most commonly seen in the United Kingdom and some parts of Europe, and has been largely replaced by the ground fault circuit interrupter (GFCI) socket. It is possible to use these types of sockets for replacement of old sockets or installation in a new bathroom design, but usually a GFCI socket is more appropriate. The newer socket style may also be required under the building code.

The bathroom shaver socket is mounted in an isolation transformer that cuts it off from the rest of the electrical system. This limits the risk of electrical shock, a concern in the bathroom where lots of water can be present. In addition, the current is very low. It will power a shaver or electric toothbrush, but not an item like a heater or a hair dryer. The circuit will shut down if it gets too hot, thus keeping the user as safe as possible.
 & In the UK, special sockets are required for bathrooms due to building regulations [2]. These sockets, called bathroom shaver sockets, are designed to be safer than a regular 3-pronged electrical socket [1]. They are mounted in an isolation transformer that cuts them off from the rest of the electrical system [1]. This limits the risk of electrical shock, which is a concern in the bathroom given that lots of water can be present [1]. The circuit will shut down if it gets too hot, which protects the user as safe as possible [1]. Additionally, these sockets are often part of a light fixture above a mirror and hidden behind a little door, which makes them harder to access [2]. & [Structure]:
 
Comparative

[Outline]:

1. Safety measures in the UK and similar regulations elsewhere (based on [1])

2. Design differences between bathroom shaver sockets and regular 3-pin plugs (based on [2])

3. Electrical safety and low-current design (based on [2])

[Answer]:

The use of shaving plugs, also known as bathroom shaver sockets, in bathrooms is primarily a result of safety regulations and design considerations that differentiate them from normal 3-pin plugs.

\#\#\# Safety Measures and Regulations

In the UK and some other countries, building regulations dictate that regular 3-pin mains power sockets are not permitted in bathrooms. This is to prevent the risk of electric shock or injury due to coming into contact with the socket while wet. Therefore, sockets are often concealed and installed at a safe distance from bathing areas.

\#\#\# Design Differences

The bathroom shaver socket is specifically designed for safe use of electrical appliances like shavers in the bathroom environment. It is characterized by its isolation transformer which segregates it from the rest of the electrical system, thereby reducing the risk of electrical shock. This design is a safety measure particularly pertinent in bathrooms where water is likely to be present.

\#\#\# Electrical Safety and Low-Current Design

The shaver socket is engineered to power only devices that are safe to use in a wet environment, such as shavers and electric toothbrushes. It is not designed to power appliances that could present a higher risk, like heaters or hair dryers. Additionally, the circuit is designed to shut down if it overheats, which further ensures user safety.

In conclusion, shaving plugs are considered safer than normal 3-pin plugs because of the specific safety regulations they adhere to, their isolated design, and the low-current circuitry that prevents riskier electrical appliances from being used in wet environments. \\ \hline
\end{longtable}

% \begin{longtable}{|p{\textwidth}|}
%     \caption{Prompt for Retrieval} \\
%     \hline
% 你的任务是从\lbrack 文档\rbrack 中抽取与\lbrack 问题\rbrack 相关且能够回答\lbrack 问题\rbrack 的内容, 请直接返回原文的片段，不要进行总结或改写。如果认为文档不能回答问题，则返回“无法回答”。\\
% 示例：\\
% \lbrack 问题\rbrack ：杭州公积金提取额度\\
% \lbrack 文档\rbrack ：2023年1月1日起，杭州无房租赁如何提取住房公积金？具体如何操作？一起来看。\\
% 今年起，无房租赁提取住房公积金限额具体标准：杭州市区（含萧山区、余杭区、临平区、富阳区、临安区）为1500元/月，桐庐县为1050元/月，淳安县为750元/月，建德市为600元/月。\\
% 温馨提示：\\
% 缴存人家庭租住公共租赁住房的，按照实际租金支出提取本人及配偶住房公积金账户余额。\\
% \lbrack 抽取结果\rbrack：无房租赁提取住房公积金限额具体标准：杭州市区（含萧山区、余杭区、临平区、富阳区、临安区）为1500元/月，桐庐县为1050元/月，淳安县为750元/月，建德市为600元/月。\\
% 请根据以上示例进行回答。\\
% \lbrack 问题\rbrack ：\textbf{\{query\}}\\
% \lbrack 文档\rbrack ：\textbf{\{paragraph\}}\\
% \lbrack 抽取结果\rbrack：\\
% \hline
% \end{longtable}

\end{CJK*}

\end{document}